%% file: main.tex
\documentclass[nohyperref]{article}

\PassOptionsToPackage{capitalise,noabbrev}{cleveref}

\usepackage[accepted,nohyperref]{include/icml2023}
\usepackage{microtype}
\usepackage{graphicx}
\usepackage{subfig}
\usepackage{booktabs} %

\usepackage{titletoc}
\usepackage{amsmath}
\usepackage{amsthm}
\usepackage{amssymb}
\usepackage{hyperref}

\usepackage{subfig}
\usepackage{multirow}
\usepackage{array}
\usepackage{multicol}
\usepackage{tabularx}
\usepackage[inline]{enumitem}

\input{include/custom}

\input{include/custom_icml2023}

\usepackage[algo2e]{algorithm2e}

\icmltitlerunning{Function-Space Regularization in Neural Networks: A Probabilistic Perspective}

\begin{document}

\twocolumn[
\icmltitle{Function-Space Regularization in Neural Networks:\\A Probabilistic Perspective}

\icmlsetsymbol{equal}{*}

\begin{icmlauthorlist}
\icmlauthor{Tim G. J. Rudner}{nyu}
\icmlauthor{Sanyam Kapoor}{nyu}
\icmlauthor{Shikai Qiu}{nyu}
\icmlauthor{Andrew Gordon Wilson}{nyu}
\end{icmlauthorlist}

\icmlaffiliation{nyu}{New York University, USA}

\icmlcorrespondingauthor{Tim G. J. Rudner}{tim.rudner@nyu.edu}
\icmlkeywords{Machine Learning, ICML}

\vskip 0.3in
]

\printAffiliationsAndNotice{}  %
\input{icml2023/0_abstract}
\input{icml2023/1_introduction}

\input{icml2023/2_background}
\input{icml2023/3_method}

\input{icml2023/5_related_work}

\input{icml2023/4_experiments}

\input{icml2023/6_conclusion}
\input{icml2023/7_acknowledgements}

\clearpage

\bibliography{references}
\bibliographystyle{include/icml2023}

\clearpage

\input{icml2023/8_appendices}

\end{document}

%% file: include/custom.tex
\usepackage[utf8]{inputenc}
\usepackage{microtype}
\usepackage{graphicx}
\usepackage{booktabs} %
\usepackage{titletoc}
\usepackage{hyperref}

\usepackage{amsmath}
\usepackage{amssymb}
\usepackage{bbm} 
\usepackage{bm}
\usepackage{verbatim}
\usepackage{float}
\usepackage{color,soul}
\usepackage{enumitem}
\usepackage{mathtools}
\usepackage{hhline}
\usepackage[title]{appendix}
\usepackage[nameinlink]{cleveref} %
\usepackage{tikz}
\usepackage{wrapfig,booktabs}
\usepackage{xspace}
\usepackage{cancel}
\usepackage{sidecap}

\graphicspath{ {../figures/} }

\DeclarePairedDelimiterX{\infdivx}[2]{(}{)}{%
  #1\;\delimsize\|\;#2%
}

\newcommand{\real}{\mathbb{R}}

\crefname{appsec}{appendix}{appendices}
\Crefname{appsec}{Appendix}{Appendices}

\definecolor{mydarkblue}{rgb}{0,0.08,0.45}
\hypersetup{ %
    pdftitle={},
    pdfauthor={},
    pdfsubject={Proceedings of the International Conference on Machine Learning 2020},
    pdfkeywords={},
    pdfborder=0 0 0,
    pdfpagemode=UseNone,
    colorlinks=true,
    linkcolor=mydarkblue,
    citecolor=mydarkblue,
    filecolor=mydarkblue,
    urlcolor=mydarkblue,
    pdfview=FitH
}

\usetikzlibrary{shapes.geometric, arrows, bayesnet, calc, positioning}

\newcommand{\dee}{\,\textrm{d}}

\newcommand{\calF}{\mathcal{F}}

\newcommand{\calN}{\mathcal{N}}

\newcommand{\calL}{\mathcal{L}}

\newcommand{\A}{\mathbf{A}}

\newcommand{\R}{\mathbb{R}}

\newcommand{\X}{\mathbf{X}}

\newcommand{\btheta}{{\boldsymbol{\theta}}}

\newcommand{\closer}[3]{{\kern-#1ex{#2}\kern-#3ex}}

\newcommand{\DKL}{D_{\text{KL}}\infdivx}

\DeclareMathOperator*{\argmax}{arg\,max}

\mathchardef\mhyphen="2D

%% file: include/custom_icml2023.tex
\usepackage{titletoc}
\usepackage{algorithm}
\usepackage{algorithmic}
\usepackage{colortbl}

\definecolor{azure}{rgb}{0.0, 0.5, 1.0}
\definecolor{airforceblue}{rgb}{0.36, 0.54, 0.66}
\definecolor{darkgreen}{rgb}{0.0, 0.2, 0.13}

\newcommand\defines{\,\dot{=}\,}

\newcommand{\map}{\textsc{map}\xspace}
\newcommand{\fsmap}{\textsc{fsmap}\xspace}

\newcommand{\vbar}{\,|\,}

\newcommand{\calX}{\mathcal{X}}
\newcommand{\calY}{\mathcal{Y}}
\newcommand{\calD}{\mathcal{D}}

\newcommand{\calQ}{\mathcal{Q}}

\newcommand{\DD}{\mathbb{D}}

\newcommand{\bXhat}{\hat{X}}

\newcommand{\bX}{X}
\newcommand{\by}{y}
\newcommand{\bY}{Y}
\newcommand{\bx}{x}

\newcommand{\pms}[1]{\ensuremath{{\scriptstyle\pm #1}}}

\usepackage{standalone}
\usepackage{tikz}
\usepackage{pgfplots}
\usepackage{pgf}
\usetikzlibrary{calc}
\usetikzlibrary{positioning}
\usetikzlibrary{angles,quotes}
\usetikzlibrary{backgrounds}
\usetikzlibrary{fit}
\usetikzlibrary{arrows}
\usetikzlibrary{arrows.meta}
\usetikzlibrary{shapes.symbols}
\usetikzlibrary{shadings}
\usetikzlibrary{shapes}
\usetikzlibrary{fadings}
\usetikzlibrary{bayesnet}
\usetikzlibrary{matrix}
\usetikzlibrary{plotmarks}
\usetikzlibrary{intersections}
\usetikzlibrary{pgfplots.fillbetween}
\pgfplotsset{compat=1.14}
\usepackage{siunitx}

\usepackage{colortbl}
\definecolor{mediumgray}{gray}{0.7}
\definecolor{lightgray}{gray}{0.85}
\definecolor{lightlightgray}{gray}{0.9}
\definecolor{C1}{HTML}{1F77B4}
\definecolor{C2}{HTML}{FF7F0E}
\definecolor{C3}{HTML}{2CA02C}
\definecolor{C4}{HTML}{D62728}
\definecolor{C5}{HTML}{9467BD}
\colorlet{C1light}{C1!70!white}
\colorlet{C2light}{C2!70!white}
\colorlet{C3light}{C3!70!white}
\colorlet{C4light}{C4!70!white}
\colorlet{C5light}{C5!70!white}
\colorlet{C1lighter}{C1!50!white}
\colorlet{C2lighter}{C2!50!white}
\colorlet{C3lighter}{C3!50!white}
\colorlet{C4lighter}{C4!50!white}
\colorlet{C5lighter}{C5!50!white}
\colorlet{C1vlight}{C1!20!white}
\colorlet{C2vlight}{C2!20!white}
\colorlet{C3vlight}{C3!20!white}
\colorlet{C4vlight}{C4!20!white}
\colorlet{C5vlight}{C5!20!white}
\colorlet{linkcolor}{violet}

\newcommand{\bnns}{\textsc{bnn}s\xspace}

\newcommand{\fsvi}{\textsc{fs-vi}\xspace}

\newcommand{\kld}{KL divergence\xspace}
\newcommand{\fsgc}{\textsc{fs-eb}\xspace}
\newcommand{\psmap}{\textsc{ps-map}\xspace}

\usepackage{tcolorbox}

\newcommand{\codebox}[2]{%
\begin{tcolorbox}[colback=blue!10!white,leftrule=2.5mm,size=title]
#1: #2
\end{tcolorbox}
\vspace{-0.1cm}%
}

%% file: icml2023/0_abstract.tex
\begin{abstract}
Parameter-space regularization in neural network optimization is a fundamental tool for improving generalization. However, standard parameter-space regularization methods make it challenging to encode explicit preferences about desired predictive \textit{functions} into neural network training. In this work, we approach regularization in neural networks from a probabilistic perspective and show that by viewing parameter-space regularization as specifying an empirical prior distribution over the model parameters, we can derive a probabilistically well-motivated regularization technique that allows explicitly encoding information about desired predictive functions into neural network training. This method---which we refer to as function-space empirical Bayes (\fsgc)---includes both parameter- and function-space regularization, is mathematically simple, easy to implement, and incurs only minimal computational overhead compared to standard regularization techniques. We evaluate the utility of this regularization technique empirically and demonstrate that the proposed method leads to near-perfect semantic shift detection, highly-calibrated predictive uncertainty estimates, successful task adaption from pre-trained models, and improved generalization under covariate shift.
\end{abstract}

%% file: icml2023/1_introduction.tex
\begin{figure}[t]
    \centering
    \begin{minipage}{0.93\columnwidth}
    \hspace*{2pt}
    \includegraphics[trim={0 0 135pt 0},clip, height=3cm]{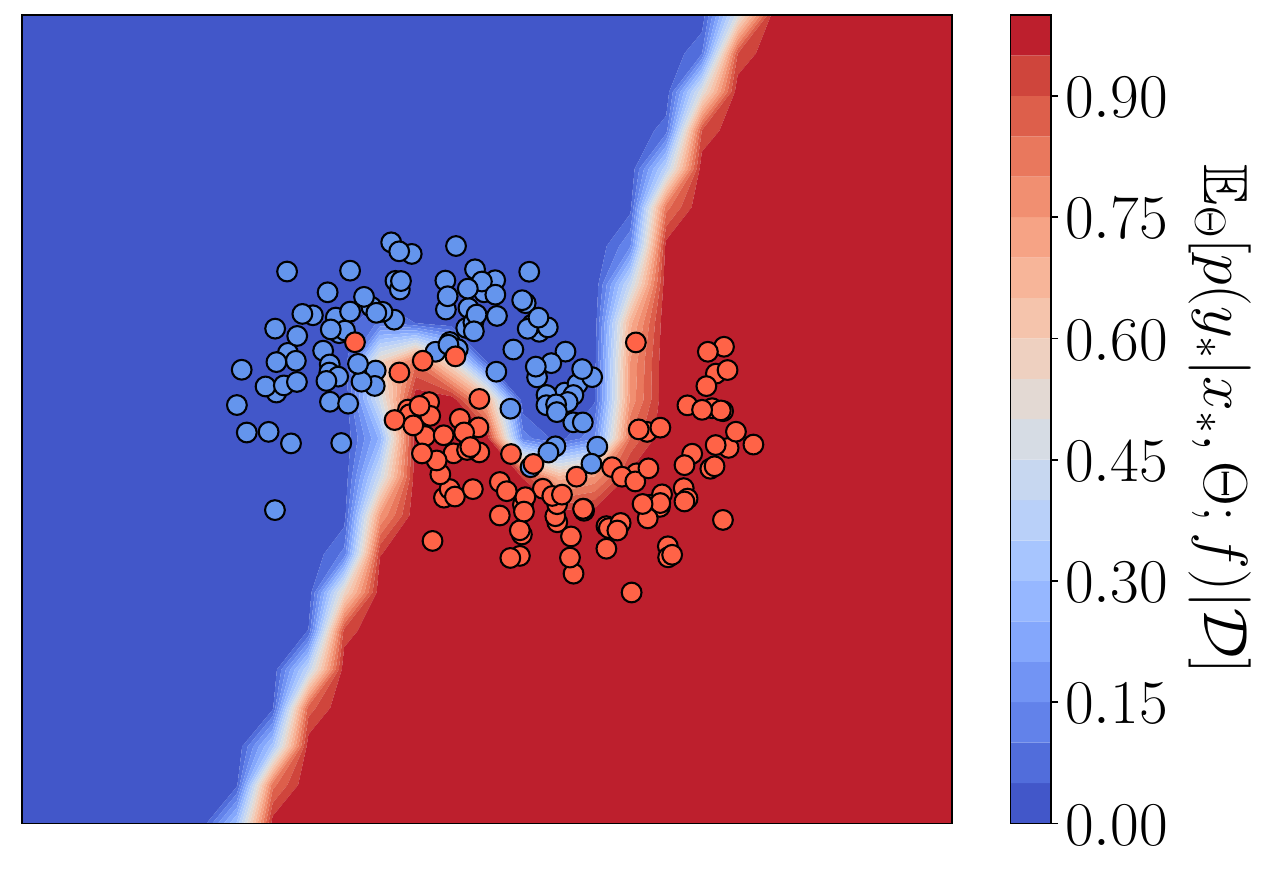}
    \includegraphics[trim={0 0 40pt 0},clip, height=3cm]{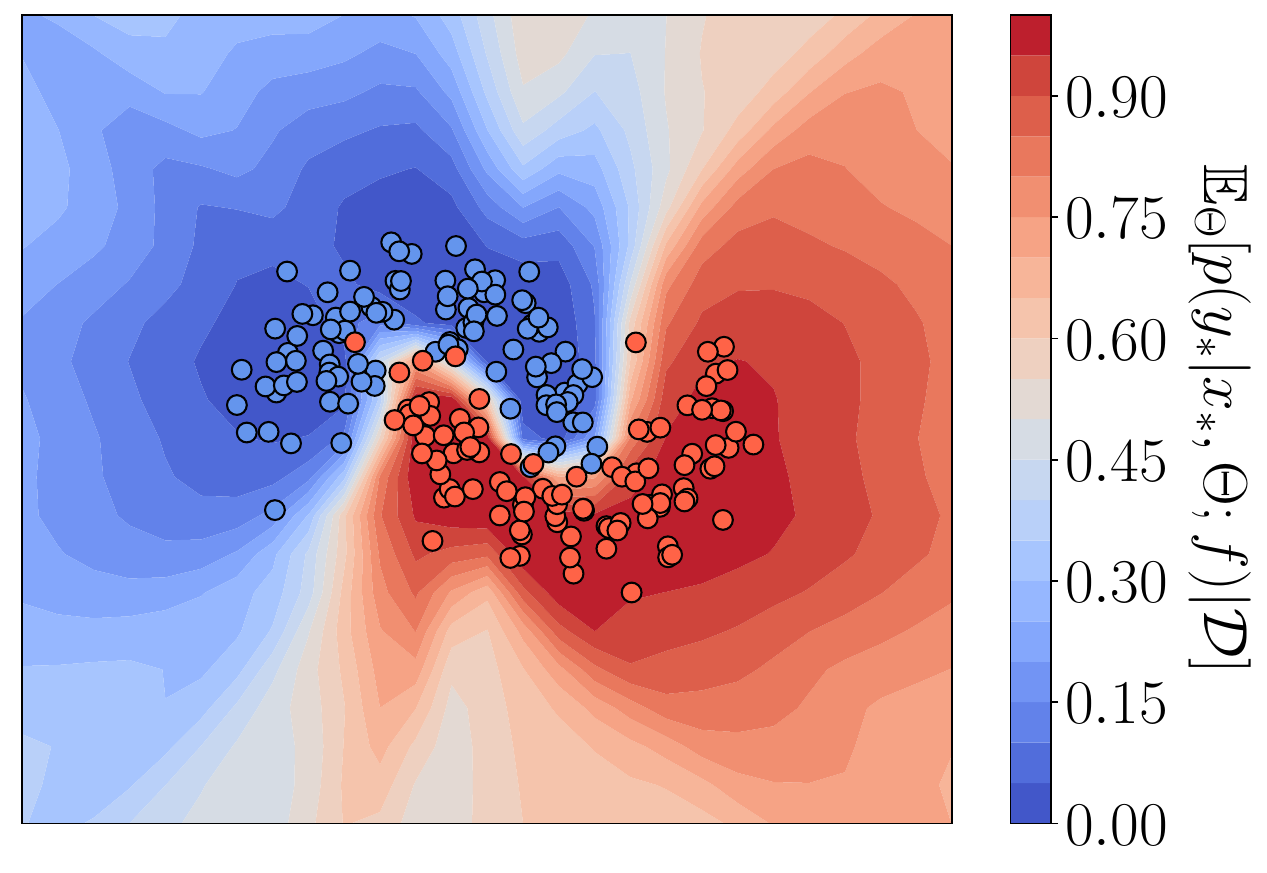}
    \\
    \hspace*{2pt}
    \includegraphics[trim={0 0 135pt 0},clip, height=3cm]{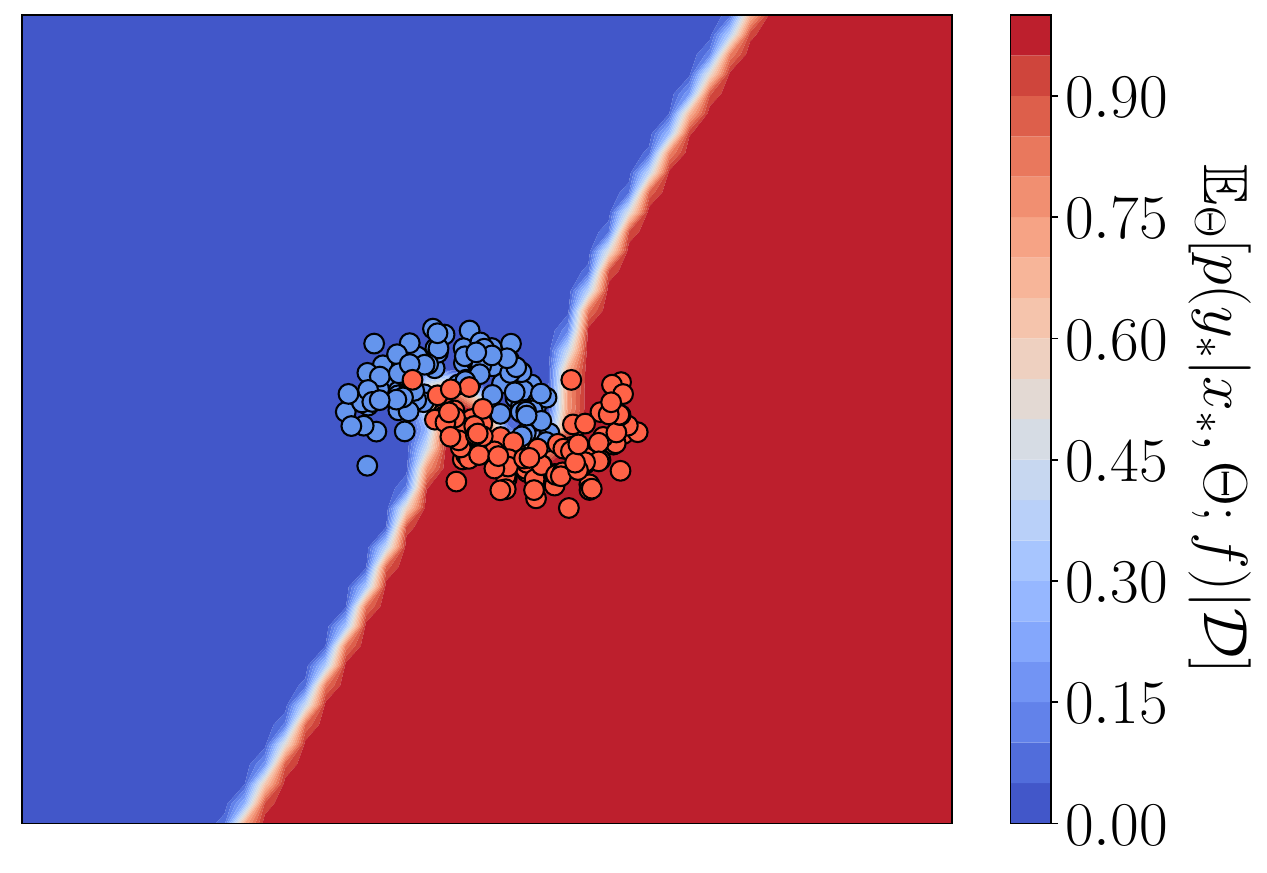}
    \includegraphics[trim={0 0 40pt 0},clip, height=3cm]{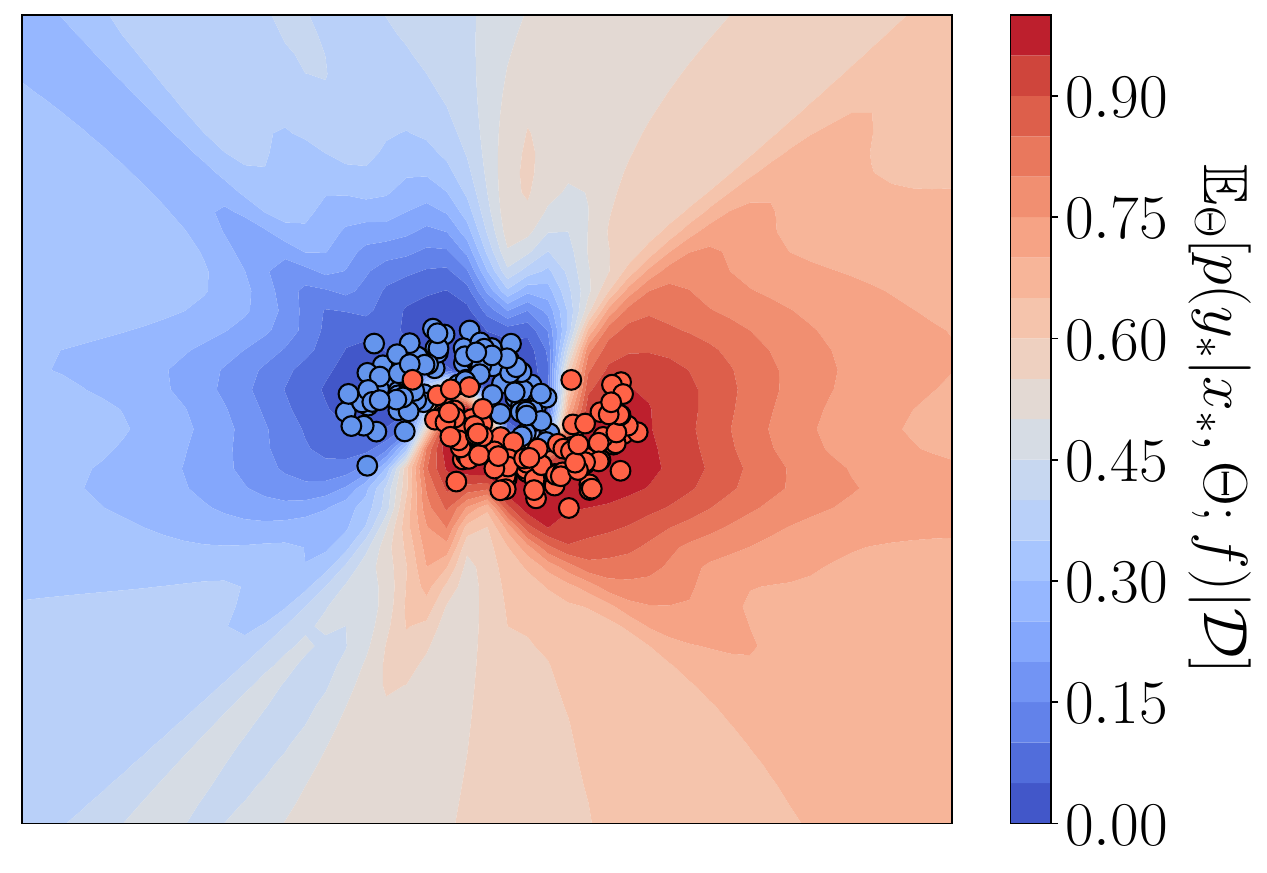}
    \end{minipage}
    \begin{minipage}{0.06\columnwidth}
        \vspace{-25pt}
        \includegraphics[trim={570pt 90pt 0 0},clip, height=3.8cm]{figures/two_moons_evi_predictive_mean.pdf}
    \end{minipage}
    \caption{
    Predictive distributions obtained by training on the \textit{Two Moons} datasets using standard parameter-space maximum a posteriori estimation (\textbf{Left}) and function-space empirical Bayes (\fsgc) (\textbf{Right}) in a two-layer MLP.
    \fsgc results in better-calibrated predictive uncertainty away from the training data, reflecting the inductive bias of the empirical prior distribution over the neural network parameters.
    }
    \label{fig:intro_figure}
\end{figure}

\section{Introduction}
\label{sec:intro}

The primary goal of machine learning is to find \textit{functions} that represent relationships in data.
Yet, most regularization methods in modern machine learning are expressed solely in terms of desired function parameters instead of the desired functions themselves.

In this work, we propose a probabilistic inference method that results in an optimization objective that features both explicit parameter- and function-space regularization.
To obtain such an optimization objective, we approach function-space regularization in deep neural networks from a probabilistic perspective and define an empirical prior distribution over parameters that allows explicitly encoding relevant prior information about the data-generating process into training.
The resulting regularizer is mathematically simple, easy to implement, and effectively induces training dynamics that encourage solutions in parameter space that are consistent with both the encoded prior information about the network parameters and the desired functions.
We refer to the probabilistic method as function-space empirical Bayes (\fsgc).

To derive an optimization objective that explicitly features parameter- and function-space regularization, we consider an empirical Bayes framework and specify an empirical prior distribution that reflects our prior beliefs about the model parameter and the predictive function induced by them.
More specifically, we consider a two-part inference problem: \begin{enumerate*}[label=(\roman*)] \item an auxiliary inference problem for finding a posterior that can be used as an empirical prior and \item a primary inference problem, where we use the empirical prior and an observation model of the data to perform Bayesian inference.\end{enumerate*}

To obtain an empirical prior that includes both parameter- and function-spaces regularizers, we consider an auxiliary inference problem, where the posterior distribution would reflect both prior beliefs about the neural network parameters (via a prior distribution over the parameters) as well as preferences about desired predictive functions (via a likelihood function that favors functions consistent with a specific distribution over functions).

We evaluate deterministic neural networks trained with the proposed regularized optimization objective on a broad range of standard classification, real-world domain adaption, and machine learning safety benchmarking tasks.
We find that the proposed method successfully biases neural network training dynamics towards solutions that reflect the inductive biases of prior distributions over neural network functions, which can yield improved predictive performance and leads to significantly improved uncertainty quantification vis-\`a-vis standard parameter-space regularization and state-of-the-art function-space regularization methods.

To summarize, our key contributions are as follows:\vspace*{-10pt}
\begin{itemize}[leftmargin=10pt]
\setlength\itemsep{0pt}
    \item
    In \Cref{sec:emp_prior_function}, we specify an auxiliary inference problem, which allows us to obtain an analytically tractable unnormalized empirical prior distribution that reflects both prior beliefs about the neural network parameters and preferences about desired predictive functions.
    \item
    In \Cref{sec:ebmap,sec:ebvi}, we show how to perform tractable maximum a posteriori estimation and approximate posterior inference in neural networks using this unnormalized empirical prior and derive an optimization objective that features both parameter- and function-spaces regularization.
    We refer to this approach as function-space empirical Bayes (\fsgc).
    \item
    In \Cref{sec:emp_eval}, we present an empirical evaluation in which we compare highly-tuned parameter- and function-space regularization baselines to neural networks trained with \fsgc regularization and find that \fsgc yields \begin{enumerate*}[label=(\roman*)] \item near-perfect semantic shift detection, \item highly-calibrated predictive uncertainty estimates, \item successful task adaption from pre-trained models, and \item improved generalization under covariate shift.\end{enumerate*}
\end{itemize}
\vspace*{-5pt}

\codebox{The code for our experiments can be accessed at}{\begin{center}
    \footnotesize
    \href{https://github.com/timrudner/function-space-empirical-bayes}{\texttt{https://github.com/timrudner/\\function-space-empirical-bayes}}.
\end{center}}
\vspace*{-20pt}

%% file: icml2023/2_background.tex
\newpage

\section{Background}

We will first review relevant background on probabilistic inference and related parameter-space and function-space regularization methods.

Consider supervised learning problems with $N$ i.i.d. data realizations \mbox{${\calD = \{\bx^{(n)}, \by^{(n)}\}_{n=1}^N} = (\mathbf{x}_\calD, \mathbf{y}_\calD)$} of inputs \mbox{$\bx \in \calX$} and targets \mbox{$\bY \in \calY$} with input space \mbox{$\calX \subseteq \real^D$} and target space \mbox{$\calY \subseteq \real^K$} for regression and \mbox{$\calY \subseteq \{0, 1\}^K$} for classification tasks with $K$ classes.

\subsection{Parameter-Space\hspace*{-1.5pt} Maximum\hspace*{-1.5pt} A\hspace*{-1.5pt} Posteriori\hspace*{-1.5pt} Estimation\hspace*{-5pt}}

For supervised learning tasks, we define a parametric observation model $p_{\bY | \bX, \Theta}(\by \vbar \bx, \theta; f)$ with mapping \mbox{$f(\cdot \,; \theta) \defines h(\cdot \,; \theta_{h}) \theta_{L}$} and a \textit{prior} distribution over the parameters, $p_{\Theta}(\theta)$.
Maximum a posteriori (\map) estimation seeks to find the most likely setting $\theta^{\map}$ of the quantity $\theta$ (under the probabilistic model) given the data.
Since, by Bayes' Theorem, the implied posterior is proportional to the joint probability density given by the product of the \textit{likelihood} of the parameters under the data $p_{\bY | \bX, \Theta}(\by_{\calD} \vbar \bx_{\calD}, \theta)$ and the prior, that is,
\begin{align*}
    p_{\Theta | \bY, \bX}(\theta \vbar \by_{\calD}, \bx_{\calD}) 
    \propto
    p_{\bY | \bX, \Theta}(\by_{\calD} \vbar \bx_{\calD} , \theta) p_{\Theta}(\theta) ,
\end{align*}
\map estimation seeks to find the mode of the joint probability density \mbox{$p(\by_{\calD} \vbar \bx_{\calD}, \theta) p(\theta)$}~\citep{Bishop2006PatternRA,murphy2013probabilistic}.
Under a likelihood that factorizes across the data points given parameters $\theta$,
\begin{align}
    p(\by_{\calD} \vbar \bx_{\calD} , \theta) \defines \prod_{n=1}^N p(\by^{(n)}_{\calD} \vbar \bx^{(n)}_{\calD} , \theta) ,
\end{align}
the \map optimization objective can be expressed as
\begin{align*}
    \calL^{\map}(\theta)
    =
    \sum_{n=1}^N \log{p_{\bY | \bX, \Theta}(\by^{(n)}_{\calD} \vbar \bx^{(n)}_{\calD} , \theta)} + \log{p_{\Theta}(\theta)} .
\end{align*}
The log-likelihood in the \map optimization objective corresponds to a scaled negative mean squared error (MSE) loss function under a Gaussian likelihood (used for regression) and to a negative cross-entropy loss function under a categorical likelihood (used for classification).

The most common instantiations of parameter-space \map estimation are $L_1$- and $L_2$-norm parameter regularization, which are also known as LASSO regression and weight decay or ridge regression, respectively.
More specifically, choosing a prior $p(\theta) = \mathcal{N}(\theta; \mathbf{0}, \sigma_0^2 I)$ leads to the standard $L_{2}$-norm regularization (also known as weight decay) and $p(\theta) = \mathrm{Laplace}(\theta; \mathbf{0}, b I)$ leads to the sparsity-inducing $L_{1}$-norm regularization (also known as LASSO)~\citep{Bishop2006PatternRA,murphy2013probabilistic}, making parameter-space \map estimation one of the most widely used optimization frameworks in modern machine learning.

\subsection{Function-Space Maximum A Posteriori Estimation}

\citet{wolpert1993fsmap} considered posterior inference over functions evaluated at a finite set of context points, $\hat{\bx} \defines \{x_{1}, ..., x_{M} \}$ to find the most likely parameters that represent the most likely function under the posterior distribution over functions.%

Letting the set of input points $\hat{\bx}$ at which the function is evaluated contain the training data such that $\bx_{\calD} \subseteq \hat{\bx}$, we can write the posterior distribution over functions at $\hat{\bx}$ as 
\begin{align*}
    p(f( \hat{\bx}) \vbar \by_{\calD}, \hat{x})
    &
    =
    p(\by_{\calD} \vbar x_{\calD}, f(\hat{\bx})) p(f(\hat{\bx}) \vbar \hat{x} ) / p(\by_{\calD} \vbar \hat{\bx} )
\end{align*}
and express the mode of the posterior via the finite-point function-space \map estimate $f( \hat{\bx}; \theta^{\fsmap})$ where $\theta^{\fsmap}$ is the mode of the finite-point function-space posterior:
\begin{align*}
\label{eq:fs_map}
    \theta^{\fsmap}
    &
    \defines
    \argmax_{\theta \in \R^{P}} p(\by_{\calD} \vbar f(\hat{\bx} ; \theta) ) p(f(\hat{\bx} ; \theta) \vbar \hat{\bx} ) .
\end{align*}
To find the finite-points function-space \textsc{map} estimate, we need to be able to maximize the joint density
\begin{align*}
    p(\by_{\calD} \vbar f(\hat{\bx} ; \theta) ) p(f(\hat{\bx} ; \theta) \vbar \hat{\bx} ) 
\end{align*}
with respect to $\theta$.
While the first term is the likelihood of the data given model parameters $\theta$, the prior density $p(f(\hat{\bx}; \theta) \vbar \hat{\bx} )$ is not in general tractable.
However, assuming that $f$ is a neural network with a standard parameterization (e.g., a multi-layer perceptron) and the set of evaluation points is sufficiently large so that $M K \geq P$, using a generalization of the change-of-variables formula,~\citet{wolpert1993fsmap} showed that the induced prior density is given by
\begin{align*}
\begin{split}
    &
    p(f(\hat{\bx} \,; \theta))
    =
    p(\theta) \, \mathrm{det}^{-1/2}(G(\theta)) ,
\end{split}
\end{align*}
where $G(\theta)$ is a $P$-by-$P$ matrix defined by
\begin{align*}
\begin{split}
    &
    G(\theta)
    \defines
    ({\partial f(\hat{\bx} \,; \theta)}/{\partial \theta})^\top ({\partial f(\hat{\bx} \,; \theta)}/{\partial \theta})
\end{split}
\end{align*}
and ${\partial f(\hat{\bx} \,; \theta)}/{\partial \theta}$ is the $MK$-by-$P$ Jacobian matrix of $f(\hat{\bx} \,; \theta)$ with respect to the parameters $\theta$.
To find $\theta^{\fsmap}$, one can maximize the log-joint density function,
\begin{align*}
\begin{split}
    &
    \log p(f( \hat{\bx} ; \theta) \vbar \by_{\calD}, \hat{\bx})
    \\
    &
    =
    \log p(\by_{\calD} \vbar f(\hat{\bx} ; \theta) ) + \log p(\theta) - \frac{1}{2} \, \log \det(G(\theta)) .
\end{split}
\end{align*}
That is, function-space \map estimation results in an optimization objective that includes parameter- and function-space regularization.
Unfortunately, computing the correction term is analytically intractable and computationally infeasible for large neural networks.
Motivated by function-space \map estimation, in \Cref{sec:ebmap}, we present an alternative probabilistic model that also features both parameter- and function-space regularization but is analytically tractable and scalable to large neural networks.

\subsection{Function-Space Variational Inference}

\label{sec:prob_f_inf}
Bayesian neural networks (\bnns) are stochastic neural networks trained using (approximate) Bayesian inference.
Denoting the parameters of such a stochastic neural network by the multivariate random variable $\Theta \in \R^P$ and letting the function mapping defined by a neural network architecture be given by \mbox{$f : \calX \times \R^P \rightarrow \R^{K}$}, then $f(\cdot \,; \Theta)$ is a random function.
For a parameter realization $\theta$, we obtain a function realization, $f(\cdot \,; \theta)$, and when evaluated at a finite collection of points $\hat{\bx} \defines \{x_{1}, ..., x_{M} \}$, $f(\hat{x} ; \Theta)$ is a multivariate random variable.

Instead of seeking to infer a posterior distribution over parameters, we may equivalently frame Bayesian inference in stochastic neural networks as inferring a posterior distribution over functions~\citep{sun2019fbnn,rudner2022tractable}.
Given a prior distribution over parameters $p(\theta)$, the probability density of the corresponding induced prior distribution over functions $p(f(\cdot))$ evaluated at a finite set of evaluation points $x$, can be expressed as
\begin{align*}
\begin{split}
    p_{F(x)}(f(x))
    =
    \int_{\R^{P}} p_{\Theta}(\theta') \, \delta( f(x ; \theta) - f(x ; \theta') ) \dee \theta' ,
\end{split}
\end{align*}
where $\delta(\cdot)$ is the Dirac delta function.
The probability density of the posterior distribution over functions $ p(f(\cdot) | \calD)$ induced by the posterior distribution over parameters $p(\theta | \calD)$, evaluated at a finite set of points, can be defined analogously and is given by
\begin{align*}
\begin{split}
    &
    p_{F(x) | \calD}(f(x) \vbar \calD)
    \\
    &
    =
    \int_{\R^{P}} p_{\Theta | \calD}(\theta' \vbar \calD) \, \delta( f(x \,; \theta) - f(x \,; \theta') ) \dee \theta' .
\end{split}
\end{align*}
Finally, defining a variational distribution over functions $q(F(\cdot))$ induced by a variational distribution over parameters $q(\theta)$,
we can frame inference over
\begin{align*}
\begin{split}
    q_{F(x)}(f(x))
    =
    \int_{\R^{P}} q_{\Theta}(\theta') \, \delta( f(x ; \theta) - f(x ; \theta') ) \dee \theta' ,
\end{split}
\end{align*}
we can frame posterior inference over stochastic functions $F(\cdot)$ variationally as
\begin{align*}
    \min_{q_{\Theta} \in \calQ} \DD_{\textrm{KL}}( q_{F( \cdot )} \,\|\, p_{F(\cdot ) | \calD} ) ,
\end{align*}
where $\calQ$ is a variational family.
Equivalently, we can express the inference problem as
\begin{align*}
    \max_{q_{\Theta} \in \calQ} \mathbb{E}_{q_{F( \cdot )}} [ \log p(y_{\calD} \vbar x_{\calD} , F( \cdot ) ) ] - \DD_{\textrm{KL}}( q_{F( \cdot )} \,\|\, p_{F(\cdot )} ) ,
\end{align*}
where $\DD_{\textrm{KL}}( q_{F( \cdot )} \,\|\, p_{F(\cdot )} )$ is an explicit regularizer on the variational distribution over functions $q(F(\cdot))$.
\citet{rudner2022tractable}, \citet{sun2019fbnn}, and \citet{Ma2021FunctionalVI} have proposed tractable approximations to this objective.
The function-space variational inference (\textsc{fs-vi}) approach by \citet{rudner2022tractable} is a state-of-the-art approximate inference method for \bnns.

%% file: icml2023/3_method.tex
\clearpage

\section{Function-Space Empirical Bayes}
\label{sec:fsgc}

Instead of considering standard, uninformative prior distributions over parameters, we consider an  empirical prior distribution over parameters, which allows us to obtain an optimization objective that combines the benefits of both standard parameter-space and explicit function-space regularization.
To obtain such an objective, we will consider a two-part inference procedure.
First, we will consider an auxiliary inference problem to derive an analytically tractable unnormalized empirical prior distribution.
We will then show how to incorporate this empirical prior into \map estimation and variational inference for the neural network parameters.
The resulting optimization objectives feature both explicit parameter- and function-space regularization.

\subsection{Empirical Priors via Distributions over Functions}
\label{sec:emp_prior_function}

We begin by specifying the auxiliary inference problem. Let \mbox{$\hat{x} = \{ x_{1}, ..., x_{M} \}$} be a set of context points with corresponding labels $\hat{y}$, and define a corresponding likelihood function $\hat{p}_{\bY | \bX, \Theta}(\hat{y} \vbar \hat{x} , \theta ; f)$ and a prior over the model parameters, $p_{\Theta}(\theta)$.
For notational simplicity, we will drop the subscripts going forward except when needed for clarity.
By Bayes' Theorem, the posterior under the context points and labels is given by
\begin{align}
    \hat{p}(\theta \vbar \hat{y}, \hat{x})
    \propto
    \hat{p}(\hat{y} \vbar \hat{x} , \theta ; f) p(\theta) .
\end{align}
To define a likelihood function that induces a posterior with desirable properties, we consider the following stochastic linear model for an arbitrary set of points $x \defines \{ x_{1}, ..., x_{M'} \}$,
\begin{align*}
    Z_{k}(\bx)
    \defines
    h(\bx ; \phi_{0}) \Psi_{k} + \varepsilon
\end{align*}\\[-15pt]
\vspace*{-19pt}
\begin{align*}
    \text{with} \quad \Psi_{k} \sim \calN(\psi ; \mu, \tau_{f}^{-1} I) \quad \text{and} \quad  \varepsilon \sim \calN(\mathbf{0}, \tau^{-1}_{f} I) ,
\end{align*}
for output dimensions $k = 1, ..., K$, where $h(\cdot \,; \phi_{0})$ is the feature mapping used to define $f$ evaluated at a set of fixed feature parameters $\phi_{0}$, $\mu$ is a set of mean parameters, and $\tau_{f}$ is a precision parameter.
This stochastic linear model induces a distribution over functions, which---when evaluated at $\hat{x}$---is given by
\begin{align*}
    \calN(z_{k}(\hat{x}) ; h(\hat{x} ; \phi_{0}) \mu_{k}, \tau_{f}^{-1} K(\hat{x}, \hat{x} ; \phi_{0}) ) ,
\end{align*}\\[-15pt]
where
\begin{align}
\SwapAboveDisplaySkip
    K(\hat{x}, \hat{x} ; \phi_{0})
    \defines
    h(\hat{x} ; \phi_{0}) h(\hat{x} ; \phi_{0})^\top + I
    \label{eq:covariance}
\end{align}
is an $M$-by-$M$ covariance matrix.
Letting $\mu = \mathbf{0}$, we obtain
\begin{align*}
    p(z_{k} \vbar \hat{x})
    =
    \calN(z_{k} ; \mathbf{0}, \tau_{f}^{-1} K(\hat{x}, \hat{x} ; \phi_{0}) ) .
    \label{eq:induced_prior_distribution}
\end{align*}
Viewing this probability density over function evaluations as a likelihood function parameterized by $\theta$, we define
\begin{align}
    \hat{p}(\hat{y}_{k} \vbar \hat{x} , \theta ; f)
    \defines
    \calN(\hat{y}_{k} ; f(\hat{x} ; \theta)_{k} , \tau_{f}^{-1} K(\hat{x}, \hat{x} ; \phi_{0}) ) ,
\end{align}
with labels $\hat{y} \defines \{\mathbf{0}, ..., \mathbf{0} \}$.
This likelihood function favors parameters $\theta$ for which $f(\hat{x} ; \theta)$ has high likelihood under the induced prior distribution over functions in \Cref{eq:induced_prior_distribution}.
Letting the likelihood factorize across output dimensions,\vspace*{-5pt}
\begin{align*}
    \hat{p}(\hat{y} \vbar \hat{x} , \theta ; f)
    \defines
    \prod_{k = 1}^{K} \hat{p}(\hat{y}_{k} \vbar \hat{x} , \theta ; f) ,
\end{align*}\\[-9pt]
defining the prior distribution over parameters as \mbox{$p(\theta) = \mathcal{N}(\theta ; \mathbf{0}, \tau^{-1}_{\theta})$}, and taking the log of the analytically tractable joint density $\hat{p}(\hat{y} \vbar \hat{x} , \theta ; f) p(\theta)$, we obtain
\begin{align*}
    &
    \log \hat{p}(\hat{y} \vbar \hat{x} , \theta ; f) + \log p(\theta)
    \\
    &
    \propto
    - \sum_{k = 1}^{K} \frac{\tau_{f}}{2} f(\hat{x} ; \theta)_{k}^\top K(\hat{x}, \hat{x} ; \phi_{0})^{-1} f(\hat{x} ; \theta)_{k} - \frac{\tau_{\theta}}{2} \|\theta\|_{2}^{2} ,
    \nonumber
\end{align*}
with proportionality up to an additive constant independent of $\theta$.
Defining\vspace*{-5pt}
\begin{align}
    \mathcal{J}(\theta, \hat{x})
    \defines
    -
    \sum_{k = 1}^{K} \frac{\tau_{f}}{2} d^{2}_{M}(f(\hat{x} ; \theta)_{k}, K(\hat{x}, \hat{x} ; \phi_{0})) 
    - 
    \frac{\tau_{\theta}}{2} \|\theta\|_{2}^{2} ,
    \label{eq:fs_map_regularizer}
\end{align}\\[-10pt]
where \mbox{$d^{2}_{M}(v, K) \defines v^\top K^{-1} v$}
is the squared Mahalanobis distance between $v$ and $\mathbf{0}$.
We therefore obtain
\vspace*{-2pt}
\begin{align*}
    \argmax_{\theta} \hat{p}(\theta \vbar \hat{y}, \hat{x})
    =
    \argmax_{\theta} \mathcal{J}(\theta, \hat{x}) .
\end{align*}\\[-7pt]
and hence, maximizing $\mathcal{J}(\theta, \hat{x})$ with respect to $\theta$ is mathematically equivalent to maximizing the posterior $\hat{p}(\theta \vbar \hat{y}, \hat{x})$ and leads to functions that are likely under the distribution over functions induced by the neural network mapping while being consistent with the prior over the network parameters.

\vspace*{-3pt}
\subsection{Empirical\hspace*{-1pt} Bayes\hspace*{-1pt} Maximum\hspace*{-1pt} A\hspace*{-1pt} Posteriori\hspace*{-1pt} Estimation}
\label{sec:ebmap}

We can now move on to the main inference problem.
Using the training data $\calD$, we wish to find a predictive function that fits the training data, generalizes well, and has well-calibrated predictive uncertainty.
To obtain such a predictive function, we will perform \map estimation using the posterior $\hat{p}(\theta \vbar \hat{y}, \hat{x})$ as an empirical prior over parameters.

Since the posterior considered above is proportional to an analytically tractable joint distribution, performing \map estimation using the posterior from the secondary inference problem as an empirical prior is straightforward.
Defining a probabilistic model with the empirical prior,\vspace*{-1pt}
\begin{align}
    p(\theta \vbar \by_{\calD}, \bx_{\calD})
    \propto
    p(\by_{\calD} \vbar \bx_{\calD} , \theta) \hat{p}(\theta \vbar \hat{y}, \hat{x}) ,
\end{align}
we can perform \map estimation by maximizing 
the empirical-\map optimization objective,
\begin{align*}
    \log p(\theta \vbar \by_{\calD}, \bx_{\calD})
    \propto
    \log p(\by_{\calD} \vbar \bx_{\calD} , \theta) + \log \hat{p}(\theta \vbar \hat{y}, \hat{x}) ,
\end{align*}
which is analytically tractable and can be expressed as\vspace*{-2pt}
\begin{align}
\label{eq:e-map-objective}
    \calL^{\textsc{eb-map}}(\theta)
    \defines
    \sum_{n=1}^N
    \hspace*{-1pt}
    \log p(\by^{(n)}_{\calD} \vbar \bx^{(n)}_{\calD} , \theta)
    \hspace*{-1pt}
    +
    \hspace*{-1pt}
    \mathcal{J}(\theta, \hat{x})
    .
\end{align}\\[-15pt]
This objective contains explicit penalties on both the parameter values (via the parameter norm $\| \theta \|_{2}^{2}$) as well as the induced function values on the set of context points (via the squared Mahalanobis distance between function evaluations and the zero vector, $ d_{M}(f(\hat{x} ; \theta)_{k}, K(\hat{x}, \hat{x} ; \phi_{0}))$).

\vspace*{-4pt}
\subsection{Empirical Bayes Variational Inference}
\label{sec:ebvi}

While the regularizer in \Cref{eq:fs_map_regularizer} may induce the desired behavior for a \textit{given} set of context points $\hat{x}$, we may instead wish to specify a \textit{distribution over context points} to cover a larger region of input space.
To obtain a tractable objective function for this setting, we consider a variational formulation of the inference problem.
Slightly changing the notation (using $\theta'$ instead of $\theta$), the probabilistic model in which we wish to perform inference---defined in terms of both the empirical prior and a prior distribution over the set of context points---is given by
\begin{align}
    p(\theta', \hat{x} \vbar \by_{\calD}, \bx_{\calD})
    \propto
    p(\by_{\calD} \vbar \bx_{\calD} , \theta') \hat{p}(\theta' \vbar \hat{y}, \hat{x}) p(\hat{x}) ,
\end{align}
with the empirical prior
\begin{align}
    \hat{p}(\theta' \vbar \hat{y}, \hat{x})
    \propto
    \hat{p}(\hat{y} \vbar \hat{x} , \theta' ; f) p(\theta') .
\end{align}
Now, defining a variational distribution
\begin{align*}
    q(\theta', \hat{x})
    \defines
    q(\theta') q(\hat{x}) ,
\end{align*}
we can frame the inference problem of finding the posterior $p(\theta', \hat{x} \vbar \by_{\calD}, \bx_{\calD})$ as a problem of optimization,
\begin{align*}
    \min_{q_{\Theta', \hat{X}} \in \calQ} \DKL{q_{\Theta', \hat{X}}}{p_{\Theta', \hat{X} | Y_{\calD}, X_{\calD}}} ,
\end{align*}
where $\calQ$ is a variational family.
If \mbox{$p_{\Theta', \hat{X} | Y_{\calD}, X_{\calD}} \in \calQ$}, then the solution to the variational minimization problem is equal to the exact posterior.
Defining $q(\hat{x}) \defines p(\hat{x})$, which further constrains the variational family, the optimization problem simplifies to
\begin{align*}
    \min_{q_{\Theta'} \in \calQ} \mathbb{E}_{p_{\hat{X}}} \left[ \DKL{q_{\Theta'}}{p_{\Theta' \vbar Y_{\calD}, X_{\calD}}} \right] ,
\end{align*}
which can equivalently be expressed as maximizing the variational objective
\begin{align*}
    \mathbb{E}_{q_{\Theta'}} [ \log p(y_{\calD} \vbar x_{\calD} , \Theta' ; f) ] - \mathbb{E}_{p_{\hat{X}}} [ \DKL{q_{\Theta'}}{p_{\Theta' \vbar \hat{Y}, \hat{X}}} ] .
\end{align*}
To obtain a tractable estimator of the regularization term, we first note that we can write
\begin{align*}
\begin{split}    
    &
    \mathbb{E}_{p_{\hat{X}}} [ \DKL{q_{\Theta'}}{p_{\Theta' \vbar \hat{Y}, \hat{X}}}] ]
    \\
    &
    =
    \mathbb{E}_{p_{\hat{X}}} [ \mathbb{E}_{q_{\Theta'}} [ \log q(\Theta') ] - \mathbb{E}_{q_{\Theta'}} [ \log p(\Theta' \vbar \hat{Y}, \hat{X}) ] ] ,
\end{split}
\end{align*}
where the first term is the negative entropy and the second term is the negative cross-entropy.
Defining a mean-field variational distribution $q(\theta') \defines \calN(\theta' ; \theta, \sigma^{2} I )$ with learnable $\theta$ and very small and fixed $\sigma^{2}$ (e.g., \mbox{$\sigma^{2} = 10^{-20}$}), the negative entropy term will be constant in $\theta$, and letting \mbox{$p(\theta) = \mathcal{N}(\theta ; \mathbf{0}, \tau^{-1}_{\theta})$} as before, we get
\begin{align*}
\begin{split}
    &
    \mathbb{E}_{p_{\hat{X}}} [ \mathbb{E}_{q_{\Theta'}} [ \log p(\Theta' \vbar \hat{Y}, \hat{X}) ] ]
    \\
    &
    \propto
    \mathbb{E}_{p_{\hat{X}}} \left[\mathbb{E}_{q_{\Theta'}} \left[ \log \hat{p}(\hat{Y} \vbar \hat{X} , \Theta' ; f) \right] + \mathbb{E}_{q_{\Theta'}} \left[ -\frac{\tau_{\theta}}{2} \|\Theta'||_{2}^{2} \right] \right] ,
\end{split}
\end{align*}
up to an additive constant independent of $\theta$.
From this expression, we can obtain an unbiased estimator of the \kld using simple Monte Carlo estimation:
\begin{align}
    \label{eq:empirical_prior_variational_monte_carlo}
    \calF(\theta)
    &
    \defines
    -
    \frac{1}{IJ} \sum_{i = 1}^{I} \sum_{j = 1}^{J} \mathcal{J}(\theta + \sigma \epsilon^{(j)}, \hat{X}^{(i)}) + C
    \\
    &
    \text{with} \quad \hat{X}^{(i)} \sim p_{\hat{X}} \quad \text{and} \quad \epsilon^{(j)} \sim \calN(\mathbf{0}, I)
    \nonumber
\end{align}
for $i=1,...,I$, $j=1,...,J$, and an additive constant $C$ independent of $\theta$.
This regularizer is an estimator of the expectation of $\mathcal{J}(\Theta, \hat{X})$ under $q_{\Theta'}$ and $p_{\hat{X}}$.
Finally, we obtain the variational objective
\begin{align}
\label{eq:e-vi-objective}
    \calL^{\textsc{eb-vi}}(\theta)
    \hspace*{-1pt}
    =
    \hspace*{-1pt}
    \frac{1}{S}
    \sum_{n=1}^N \sum_{s = 1}^{S}  \log p(\by^{(n)}_{\calD} \vbar \bx^{(n)}_{\calD} , \theta + \sigma \epsilon^{(s)}) - \calF(\theta) ,
\end{align}\\[-10pt]
with $\epsilon^{(s)} \sim \calN(\mathbf{0}, I)$.
This objective factorizes across training data points and, as such, is amenable to stochastic gradient descent.
This objective is used in the empirical evaluation in \Cref{sec:emp_eval}.
We will refer to this method as function-space empirical Bayes (\fsgc).

\vspace*{-4pt}
\subsection{Function-Space\hspace*{-1pt} Regularization\hspace*{-1pt} via\hspace*{-1pt} Empirical\hspace*{-1pt} Priors}

The tractable empirical-Bayes \map estimation and variational inference objectives in \Cref{eq:e-map-objective,eq:e-vi-objective}, respectively, are both defined in terms of the empirical-Bayes regularizer $J(\theta, \hat{x})$ given in \Cref{eq:fs_map_regularizer}.

First, unlike function-space regularizers proposed in prior work~\citep[e.g.,][]{bietti2019kernel,Benjamin2018MeasuringAR,sun2019fbnn,rudner2022tractable,Rudner2022sfsvi,chen2022ntk}, the regularizer $\mathcal{J}(\theta, \hat{x})$, explicitly features parameter-space regularization.
Prior distributions over parameters, such as isotropic Gaussians or the Laplace distribution, are well-established and have been demonstrated to yield parameter \map estimates that define predictive functions that generalize well.
Second, via the labels $\hat{y} = \{ \mathbf{0}, ..., \mathbf{0} \}$ used in the likelihood function, the parameters $\theta$ are encouraged to be concentrated around values that fit the training data and are consistent with both the prior distribution over parameters---which favors parameters $\theta$ with small norm $\| \theta \|_{2}^{2}$---and the likelihood function---which favors parameters $\theta$ that produce zero predictions, corresponding to high-entropy predictive distributions in classification settings and a reversion to the data mean in regression settings with normalized data.
Third, for non-singleton sets of context points $\hat{x}$, the likelihood function enforces a smoothness constraint via its covariance matrix and encourages parameters that induce functions that have high likelihood under the induced distribution over functions defined \Cref{eq:induced_prior_distribution}---which has been shown introduce desirable inductive biases into the learned model~\cite{wilson2020Bayesian,rudner2022tractable,Rudner2022sfsvi}.

\vspace*{-4pt}
\subsection{Specifying\hspace*{-1pt} Distributions\hspace*{-1pt} over\hspace*{-1pt} Sets\hspace*{-1pt} of\hspace*{-1pt} Context\hspace*{-1pt} Points}

Careful specification of $p_{\bXhat}$ is crucial for ensuring that the empirical-Bayes regularizer effectively encourages desired properties in the learned predictive functions.
A simple approach to specifying $p_{\bXhat}$ is to define the context distribution as an empirical distribution given by a dataset that is meaningfully related to the training data.
For example, we may choose an unaltered subset of the training data, corruptions/augmentations of the training data (using standard augmentations such as cropping, blurring, pixelation, etc.), or a related dataset, such as KMNIST when training on FashionMNIST or CIFAR-100 when training on CIFAR-10, as the context distribution.
In principle, the more the most relevant regions of a given problem-specific input space (e.g., the space of natural images for general image classification) are covered by a context distribution $p_{\bXhat}$, the more likely the learned function will be drawn towards the prior distribution over functions evaluated at these parts of input space.

\vspace*{-4pt}
\subsection{Specifying Prior Distributions over Functions}

When a pretrained model is available, a likelihood $\hat{p}(\hat{y} \vbar \hat{x}, \theta; f)$ can be constructed from a prior distribution over functions by specifying $\phi_{0}$ in $h(\hat{x}; \phi_{0})$ to be the pretrained model parameters.
If a pretrained model is unavailable, $\phi_{0}$ can be specified by randomly initializing the network parameters using any standard initialization scheme, which also induces desirable inductive biases \citep{wilson2020Bayesian}.

%% file: icml2023/5_related_work.tex
\vspace*{-4pt}
\section{Related Work}
\label{sec:related_work}

\citet{Krogh1991ASW} argued that explicit regularization via weight decay, that is, an $L_{2}$-norm penalty on the parameters, can significantly improve generalization.
This approach is now standard practice for training parametric models, including large neural networks.
Weight decay corresponds to maximum a posteriori estimation in probabilistic models with a Gaussian prior distribution over the model parameters.
\citet{Joo2020DeepLR} further demonstrated the effectiveness of explicit regularization for calibration of neural networks. Our work takes this case further by regularizing directly in the function space.

\citet{wolpert1993fsmap} argued that the true goal of maximum a posteriori estimation in parametric models---and, as such, of parameter-space regularization---is to find the most likely function mapping that describes the given data and the prior while the parameter-space representation of the network is only a means to an end.
However, in non-linear parametric models, since maximum a posteriori estimation is not invariant under parameterization, the function implied by the most likely parameters can differ significantly from the most probable function~\citep{Denker1990TransformingNO}.
Using the generalized change-of-variables formula for probability distributions to get the implied distribution over functions from the distribution over parameters, \citet{wolpert1993fsmap} introduced a correction term to standard parameter-space regularization with weight decay limited to small neural networks.
In contrast, we provide an alternative model formulation that leads to tractable function-space regularization for any neural network architecture.

\citet{Wang2019FunctionSP} reasoned why a good approximation to the parameter-space posterior does not necessarily correspond to better predictive performance because of symmetries in overparameterized neural networks.
Empirically,~\citet{Joo2020DeepLR} provided evidence that $L_{p}$ norm regularization in function space improves generalization in neural network models while also improving calibration.
\citet{bietti2019kernel} proposed to use the Jacobian norm as a lower bound on the function norm and~\citet{bietti2018group} constructed an RKHS which contains CNN prediction functions.
\citet{chen2022ntk} use a Mahalanobis distance regularizer between logits, with the covariance matrix given by the empirical neural tangent kernel.
In this work, we instead take an empirical Bayes approach to derive a function-space regularization objective from inference in a probabilistic model of the data-generating process.

In the context of approximate Bayesian inference, \citet{Sun2019FunctionalVB} proposed to minimize the divergence between two distributions over functions via a function-space evidence lower bound (ELBO), but \citet{Burt2020UnderstandingVI} showed that the inference problem as considered in \citet{sun2019fbnn} is not well-defined for neural network variational distributions with Gaussian process priors.
Other approaches to approximate function-space inference have been proposed \citep{Ma2018VariationalIP, Ober2020GlobalIP, Ma2021FunctionalVI}.
By instead linearizing the function mapping to obtain a tractable distribution over functions, \citet{rudner2022tractable} introduced an effective and scalable approximation to make function-space variational inference effective and scalable to deep neural networks.
\citet{Titsias2019FunctionalRF} applied functional regularization using Gaussian process priors to handle catastrophic forgetting in continual learning and~\citet{Rudner2022sfsvi} use function-space variational inference to prevent catastrophic forgetting by encouraging neural networks to match an empirical prior distribution over functions. We reiterate, however, that our work does not aim to propose a new approximate Bayesian inference approach. Instead, we investigate the utility of approximate inference with a function-space regularizer specified via empirical Bayes on the parameters.

%% file: icml2023/4_experiments.tex
\section{Empirical Evaluation}
\label{sec:emp_eval}

In this section, we evaluate empirical variational inference (\fsgc) along various dimensions---generalization (accuracy), uncertainty quantification (selective prediction, calibration), robustness (semantic shift detection, generalization under covariate shift), and transfer learning.

\textbf{Overview.}\quad
We assess whether \fsgc can improve the reliability of neural networks.
We put a special emphasis on benchmarking tasks and evaluation metrics that assess reliability as a function of predictive accuracy and predictive uncertainty estimates. 
Across all benchmarking tasks, we find that \fsgc results in improved predictive uncertainty, evaluated in terms of $\log$-likelihood, expected calibration error (ECE), and selective prediction when compared to standard parameter-space \map (denoted by \psmap).
Notably, we achieve \emph{near-perfect} semantic shift detection on both CIFAR-10 and FashionMNIST against samples from datasets that were unseen during training and do not belong to the same distribution.
We further demonstrate that \fsgc can often improve robustness to corruptions compared to parameter-space inference.

\textbf{Illustrative Example.}\quad
In \Cref{fig:intro_figure}, we illustrate the effect of \fsgc on the \textit{Two Moons} classification dataset.
On one hand, a standard data fit using standard parameter-space \map estimation shows that the model learns a decision boundary which roughly splits the space into two regions within which the model makes predictions with very high confidence.
\fsgc, on the other hand, exhibits an increase in predictive uncertainty in regions further away from the training data, where it encourages the neural network to match the prior distribution over functions (via the empirical prior), providing a more reliable solution that aligns with our a priori desire of lower confidence predictions in regions of input space far away from the training data.

\textbf{Setup.}\quad
All of our methods are trained using a ResNet-18 architecture \citep{he2016deep} with momentum SGD.
All results are reported with mean and standard error over five trials.
See \Cref{appsec:experiments} for details about hyperparameters.

\textbf{Implementation.}\quad
The optimization objective in \Cref{eq:e-vi-objective} can be implemented on top of standard training routines.
It only requires the neural network feature $h(\hat{x} ; \phi_{0})$ and the predictions $f(\hat{x} ; \theta)$ for a given sample of context points $\hat{x}$.
In practice, we use only a single Monte Carlo sample per gradient step, that is, $I = J = 1$.

\subsection{Selective Prediction}
\label{sec:selective_prediction}

Selective prediction modifies the standard prediction pipeline by introducing a ``reject option'', $\perp$, via a gating mechanism defined by a selection function $s: \calX \rightarrow \mathbb{R}$ that determines whether a prediction should be made for a given input point $\bx \in \calX$~\citep{elyaniv2010foundations,rabanser2022selective}.
For a rejection threshold $\tau$, the prediction model is then given by
\begin{align}
\SwapAboveDisplaySkip
    (p(\by \vbar \cdot, \btheta ; f), s)(\bx)
    =
    \begin{cases} 
      p(\by \vbar \bx, \btheta ; f) & s \leq \tau \\
      \perp & \text{otherwise} .
    \end{cases}
\end{align}
To evaluate the predictive performance of a prediction model $(p(\by \vbar \cdot, \btheta ; f), s)(\bx)$, we compute the predictive performance of the classifier $p(\by \vbar \bx, \btheta ; f)$ over a range of thresholds $\tau$, and summarize as the area under the selective prediction accuracy curve.
Successful selective prediction models obtain high cumulative accuracy over many thresholds and can be applied in safety-critical real-world tasks where uncertainty-aware predictive accuracy is especially important.

\begin{figure*}[!ht]
    \centering
    \begin{tabular}{c}
    \includegraphics[width=.43\linewidth]{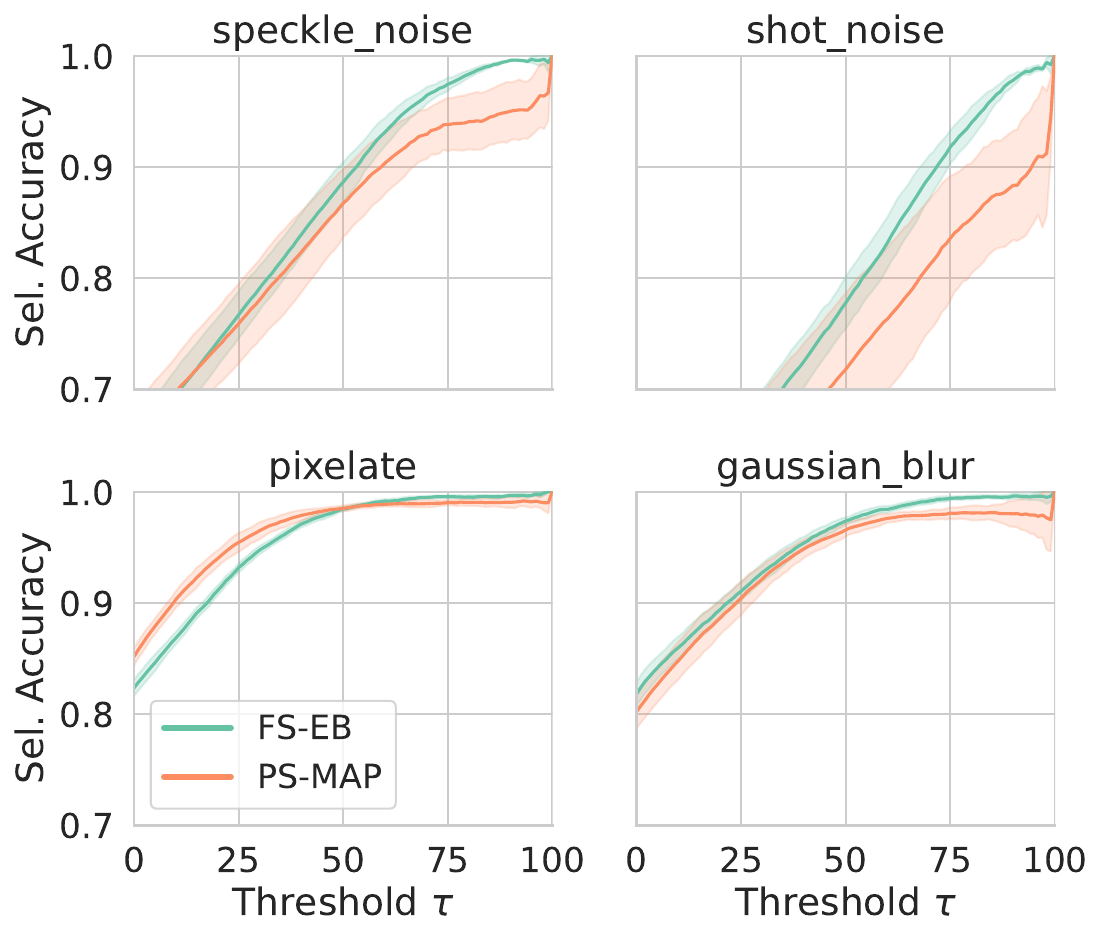}
    \hspace*{20pt}
    \includegraphics[width=.43\linewidth]{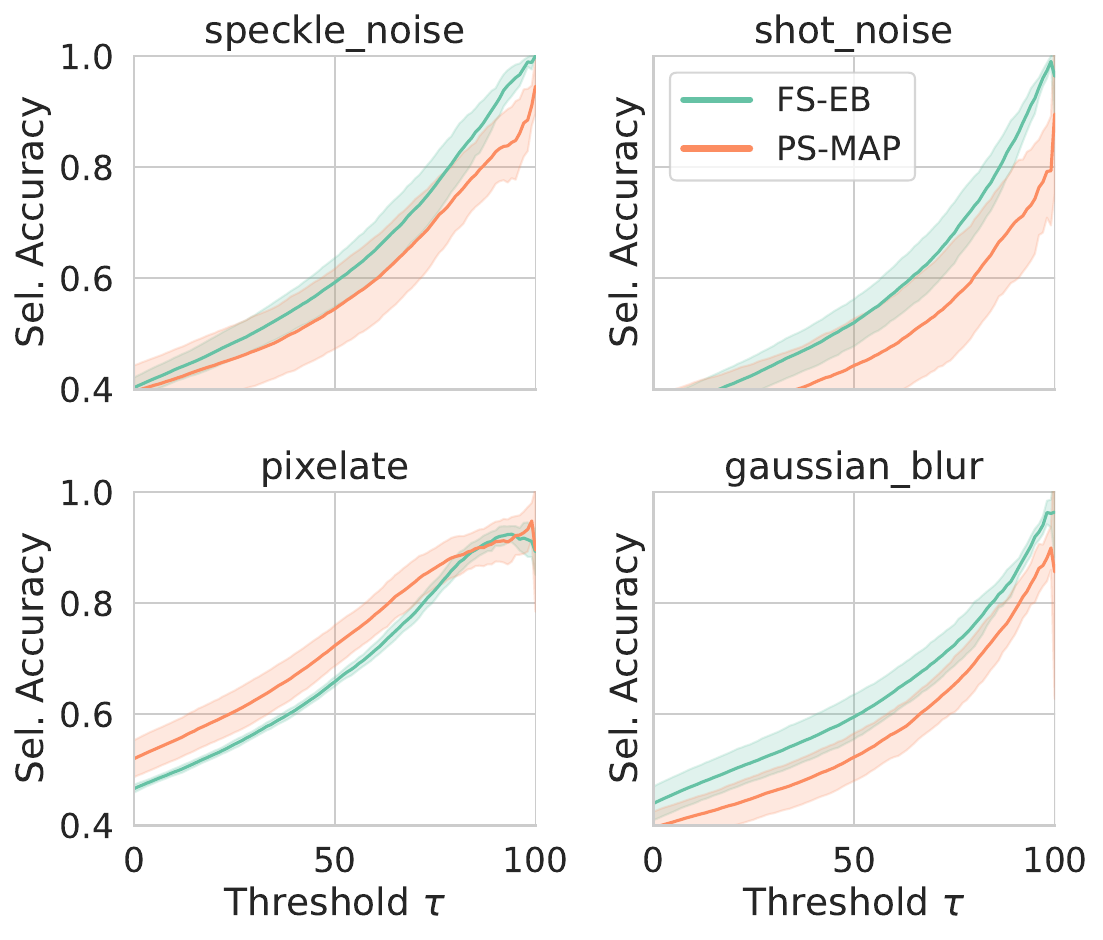}
    \\
     \centering
    (a) Corruption Level 3
    \hspace*{120pt}
    (b) Corruption Level 5
    \end{tabular}
    \caption{For a selected subset of corruptions as constructed by Corrupted CIFAR-10 \citep{hendrycks2018benchmarking}, we show the selective prediction curves for \textbf{(a)} corruption level 3 and \textbf{(b)} corruption level 5. A higher curve indicates better calibration for ``reject option'' in classification \citep{elyaniv2010foundations}. We find that \fsgc is often better out-of-the-box for certain standard image blurring and noise corruptions, indicating better calibration when compared to standard \psmap.}
    \label{fig:c10c_sel_pred}
    \vspace*{-8pt}
\end{figure*}

\Cref{fig:c10c_sel_pred} shows that \fsgc can often provide better out-of-the-box for certain standard image corruptions, tested on the Corrupted CIFAR-10 \citep{hendrycks2018benchmarking} dataset.
We plot the selective prediction accuracy curves, that is, accuracy versus confidence, such that below a chosen confidence level $\tau$, the sample is not being classified. 
Additionally, in \Cref{tab:fmnist_unc,tab:c10_unc}, we see that \fsgc improves the area under selective prediction curves, while improving the generalization of the classifier as measured by accuracy.
In practice, a fraction $1-\tau$ of the samples could get referred to a human expert for manual review.
The area under the selective prediction accuracy curves, therefore, provides information about the reliability of a classifier.

\setlength{\tabcolsep}{1.4pt}
\begin{table*}[ht]
\vspace*{-3pt}
\begin{minipage}{\columnwidth}
  \caption{
      We report the accuracy ({\sc acc.}), negative log-likelihood ({\sc nll}), expected calibration error ({\sc ece}), and area under the selective prediction accuracy curve ({\sc Sel. Pred.}) for FashionMNIST \citep{xiao2017/online} and \fsgc improves performance while improving calibration. \mbox{$\bx_\mathrm{C} = \mathrm{KMNIST}$}. Means and standard errors are computed over five seeds.
  }
  \label{tab:fmnist_unc}
  \vspace*{-7pt}
  \begin{sc}
  \begin{tabular}{l|cccc}
  \toprule
  Method & Acc. $\uparrow$ & Sel. Pred. $\uparrow$ & NLL $\downarrow$ & ECE $\downarrow$ \\
  \midrule
  \psmap & $93.8\% \pms{0.0}$ & \cellcolor[gray]{0.9}$\mathbf{98.9}\% \pms{0.0}$ & $0.26 \pms{0.00}$ & $3.6\% \pms{0.0}$ \\
  \fsgc & \cellcolor[gray]{0.9}$\mathbf{94.1}\% \pms{0.1}$ & $98.8\% \pms{0.0}$ & \cellcolor[gray]{0.9}$\mathbf{0.19} \pms{0.00}$ & \cellcolor[gray]{0.9}$\mathbf{1.8}\% \pms{0.1}$ \\
  \textsc{fs-vi} & \cellcolor[gray]{0.9}$\mathbf{94.1}\% \pms{0.0}$ & $98.4\% \pms{0.0}$ & $0.24 \pms{0.00}$ & $2.6\% \pms{0.1}$ \\
  \bottomrule
  \end{tabular}
  \end{sc}
\end{minipage}
\hfill \vrule height 2.4cm depth 2.2cm width .4pt \hfill
\setlength{\tabcolsep}{1.4pt}
\begin{minipage}{\columnwidth}
  \caption{
      We report the accuracy ({\sc acc.}), negative log-likelihood ({\sc nll}), expected calibration error ({\sc ece}), and area under the selective prediction accuracy curve ({\sc Sel. Pred.}) for CIFAR-10 \citep{Krizhevsky2010ConvolutionalDB} and \fsgc improves predictive performance and calibration. \mbox{$\bx_\mathrm{C} = \mathrm{CIFAR\text{-}100}$}. Means and standard errors are computed over five seeds.
  }
  \label{tab:c10_unc}
  \vspace*{-7pt}
  \begin{sc}
  \begin{tabular}{l|cccc}
  \toprule
  Method & Acc. $\uparrow$ & Sel. Pred. $\uparrow$ & NLL $\downarrow$ & ECE $\downarrow$ \\
  \midrule
  \psmap & $94.9\% \pms{0.2}$ & $99.3\% \pms{0.0}$ & $0.21 \pms{0.01}$ & $3.0\% \pms{0.1}$ \\
  \fsgc & \cellcolor[gray]{0.9}$\mathbf{95.1}\% \pms{0.1}$ & \cellcolor[gray]{0.9}$\mathbf{99.4}\% \pms{0.0}$ & \cellcolor[gray]{0.9}$\mathbf{0.20} \pms{0.00}$ & \cellcolor[gray]{0.9}$\mathbf{2.1}\% \pms{0.1}$ \\
  \textsc{fs-vi} & $92.9\% \pms{0.1}$ & $98.0\% \pms{0.0}$ & $0.31 \pms{0.00}$ & $4.0\% \pms{0.1}$ \\
  \bottomrule
  \end{tabular}
  \end{sc}
\end{minipage}
\end{table*}

\subsection{Calibrated Predictive Uncertainty}

As shown in \Cref{fig:intro_figure}, \psmap tends to be very confident even far away from data. 
Such predictive behavior may often be undesirable.
The expected calibration error (ECE; \citet{Naeini2015ObtainingWC}) computes the alignment between accuracy and prediction of a classifier.
In line with our illustration, through our benchmark experiments, we provide evidence that \fsgc is able to significantly improve classification calibration.

Following \citet{Naeini2015ObtainingWC}, an empirical ECE estimator is constructed by binning the maximum output probability of each sample into $m$ bins $B_j ~\forall ~ j \in [1,\dots,m]$, such that
\begin{align}
    \widehat{\text{ECE}} = \sum_{i=1}^n \frac{B_i}{n} \left| \mathrm{Accuracy}(B_i) - \mathrm{Confidence}(B_i)  \right|,
\end{align}
where $\mathrm{Acc.}$ is the accuracy of each sample within each bin $B_i$, and $\mathrm{Conf.}$ is the mean of all maximum probability outputs of a classifier for each sample within the bin $B_j$.
Therefore, a perfectly calibrated model has an ECE of zero, implying perfect alignment between the accuracy of the classifier and its confidence in the predictions.

In \Cref{tab:fmnist_unc,tab:c10_unc}, we verify that \fsgc significantly improves calibration while improving the generalization of the classifier as measured by accuracy.

\subsection{Highly-Accurate Semantic Shift Detection}

So far, we have demonstrated that \fsgc can improve the quality of neural networks' predictive uncertainty on in-domain data. 
Another hallmark of a reliable model is its ability to detect semantic shifts in the data~\citep{Band2021benchmarking,Nado2021UncertaintyBaselines}.
We assess whether the \fsgc generates predictive uncertainty estimates that enable successful semantic shift detection, that is, detection of input points whose true labels are semantically different from the training labels, and find that \fsgc can achieve \emph{near-perfect} semantic shift detection in two image classification tasks.
To simulate semantic shift, we present a classifier trained on FashionMNIST \citep{xiao2017/online}, a grayscale collection of fashion items to distinguish against KMNIST \citep{Clanuwat2018DeepLF}, with a dataset of handwritten Kuzushiji digits.

\begin{figure*}[!ht]
\vspace*{-5pt}
    \centering
    \subfloat[Accuracy]{
    \includegraphics[width=.42\linewidth]{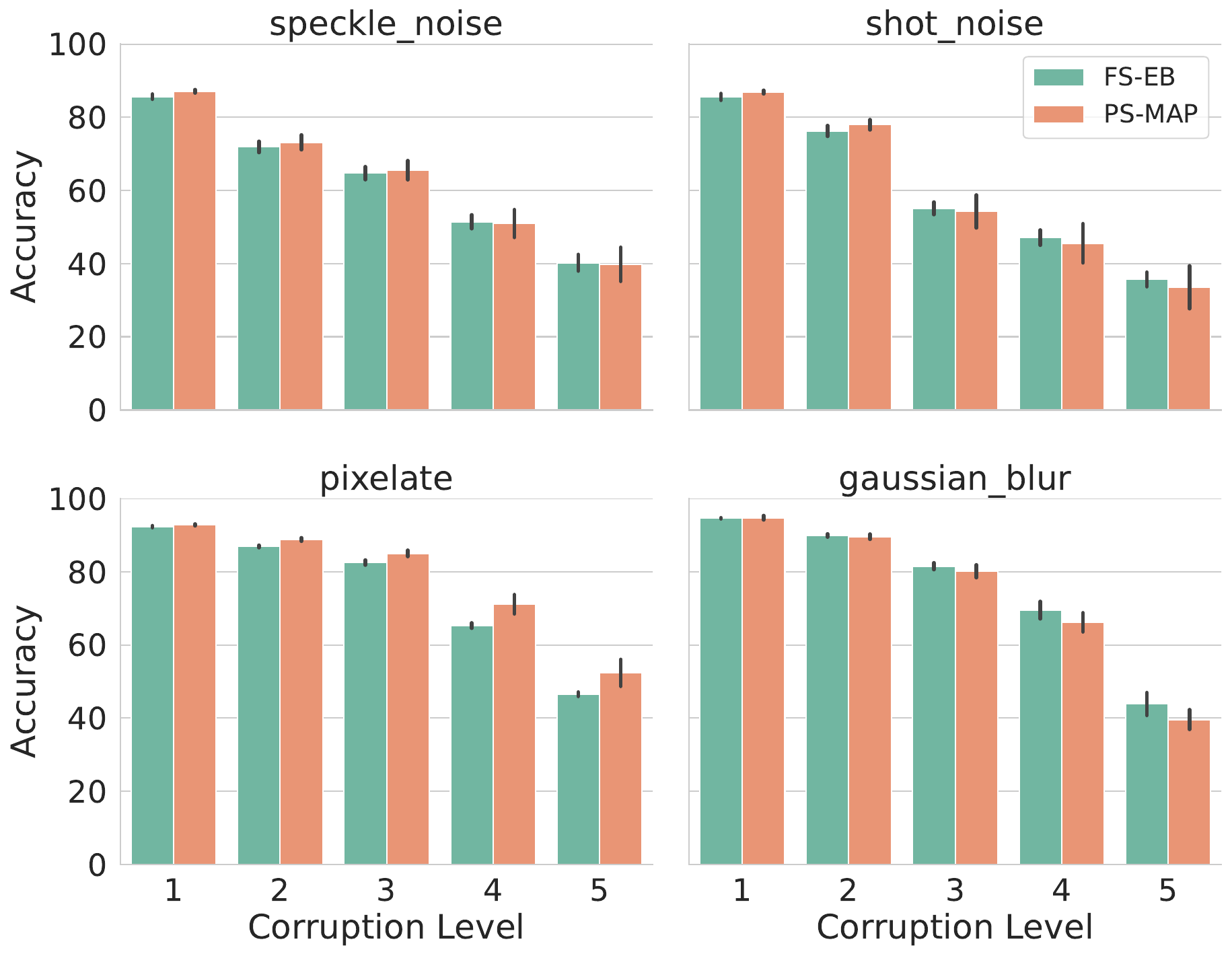}}
    \hspace*{30pt}\subfloat[Selective Prediction AUC]{
    \includegraphics[width=.42\linewidth]{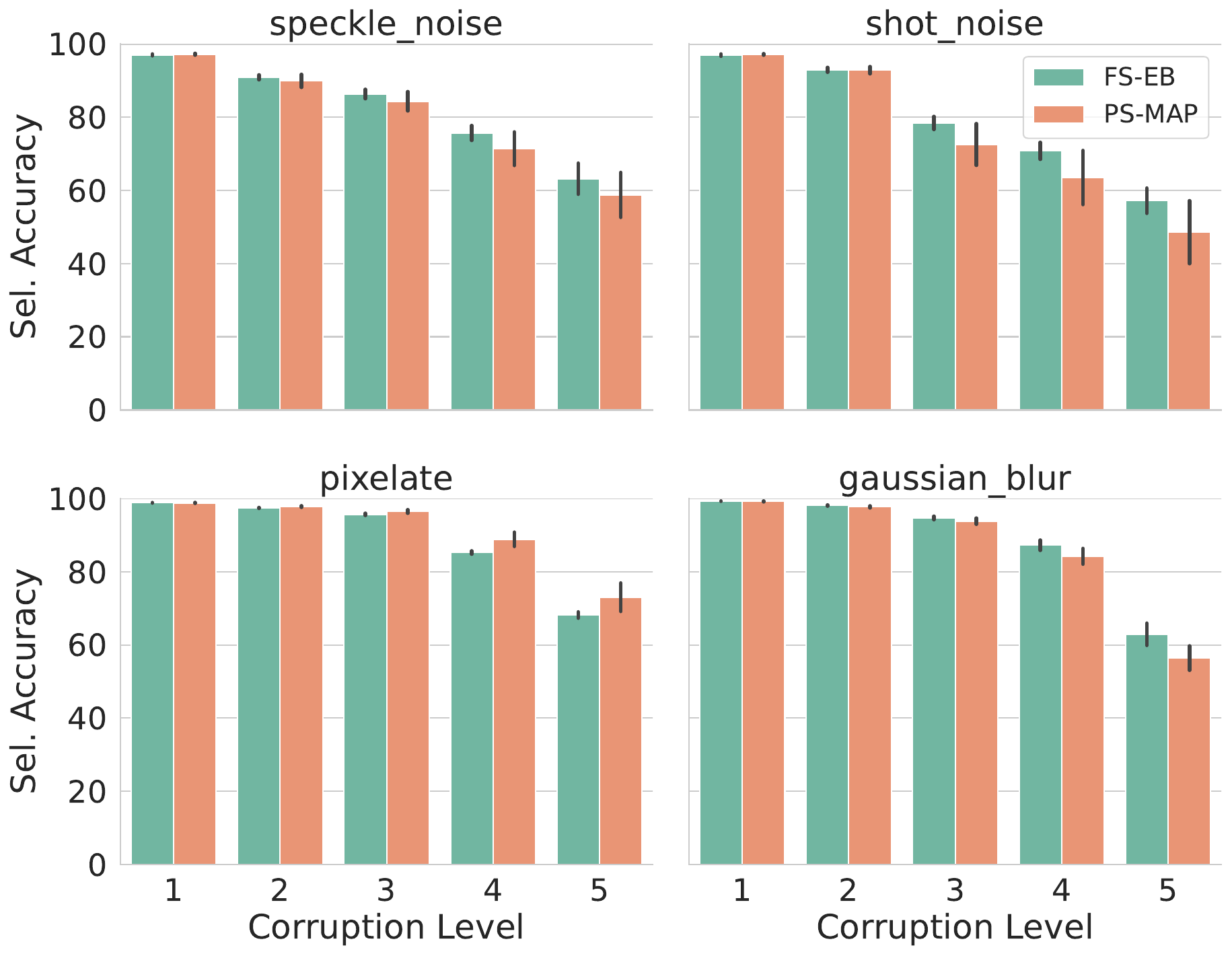}
    }
    \vspace*{-3pt}
    \caption{
        For a randomly selected subset of corruptions as constructed by Corrupted CIFAR-10 \citep{hendrycks2018benchmarking}, we show that \textbf{(a)} \fsgc and \psmap achieve similar predictive accuracy, but \textbf{(b)} \fsgc leads to better selective prediction (as measured by the area under the selective prediction accuracy curve).
        The improvement in selective prediction indicates that \fsgc produces more accurate uncertainty estimates and is thus able to use the ``reject option'' more effectively, leading to more reliable classification.
        See \Cref{fig:c10_acc_all,fig:c10_sel_acc_all} in \Cref{appsec:experiments} for results on other common corruptions.
        }
    \label{fig:c10c_acc}
    \vspace*{-10pt}
\end{figure*}

Using the predictive entropy of the classifier for each input sample from both FashionMNIST and KMNIST, we build another binary classifier to detect semantic shifts using simply the threshold of predictive entropy.
We are able to detect semantic shift with $\emph{near-perfect}$ accuracy of $99.9\%$. We come to a similar conclusion when detecting semantic shift between CIFAR-10 \citep{Krizhevsky2010ConvolutionalDB}, a collection of tiny images of objects and SVHN \citep{Netzer2011ReadingDI}, a collection of street view house numbers. Numerical results are summarized in \Cref{tab:all_ood}.

\setlength{\tabcolsep}{1pt}
\begin{table}[ht]
\vspace*{-5pt}
\caption{
    We compute the area under the ROC of a classifier using the predictive entropy on the in-distribution samples and out-of-distribution samples $\bx_{\mathrm{OOD}}$ with semantic shift. For FashionMNIST, we use ${\bx_{\mathrm{OOD}} = \text{\sc MNIST}}$; for CIFAR-10, we use ${\bx_{\mathrm{OOD}}= \text{\sc SVHN}}$.
    }
\label{tab:all_ood}
\vspace*{-14pt}
\begin{center}
\begin{sc}
\begin{tabular}{l|lc}
\toprule
\multicolumn{1}{p{0.7in}|}{Dataset} & \hspace*{4pt}Method & OOD AUROC $\uparrow$ \\
\midrule
\multirow{3}{*}{FMNIST} & \psmap & $94.9\% \pms{0.4}$ \\
& \fsgc ($\bx_C$ = KMNIST) & \cellcolor[gray]{0.9}$\mathbf{99.9}\% \pms{0.0}$ \\
& \fsvi & $98.0\% \pms{0.4}$ \\
\midrule
\multirow{3}{*}{CIFAR-10} & \psmap & $93.0\% \pms{0.4}$ \\
& \fsgc ($\bx_C$ = CIFAR100)\hspace*{2pt} & \cellcolor[gray]{0.9}$\mathbf{99.4}\% \pms{0.1}$ \\
& \fsvi & $99.0\% \pms{0.1}$ \\
\bottomrule
\end{tabular}
\end{sc}
\end{center}
\vspace*{-10pt}
\end{table}

\vspace*{-3pt}
\subsection{Generalization under Covariate Shift}

Another essential property of a reliable classifier is graceful degradation under covariate shift.
We assess the performance of \fsgc in terms of generalization under covariate shift.
Using the CIFAR-10 Corrupted dataset \citep{hendrycks2018benchmarking} at five different corruption intensity levels, we find that \fsgc can still generalize well.
In \Cref{fig:c10c_acc}, we find that \fsgc often works out-of-the-box for generalization under common visual corruptions.

\vspace*{-3pt}
\subsection{Improved Transfer Learning}

In addition to training from scratch, we also investigate the utility of \fsgc for transfer learning, a paradigm that is now very common with the advent of large pretrained neural network models \citep{Brown2020LanguageMA,Radford2021LearningTV,Tran2021plex,Touvron2023LLaMAOA}.

We find that \fsgc improves uncertainty quantification of transfer-learned models without compromising predictive performance.
\Cref{tab:c10_transfer} shows that \fsgc and \psmap reach the same level of accuracy and selective prediction AUC, but \fsgc significantly improves NLL, calibration as measured by ECE, and effective semantic shift detection, using a ResNet-18 \citep{he2016deep} pretrained on ImageNet \citep{Russakovsky2014ImageNetLS}.

In addition, we evaluate transfer-learned classifiers with \fsgc on real-world datasets. 
Using a ResNet-50 pretrained on ImageNet, we train classifiers on blindness detection, leaf disease classification, and melanoma detection and find that \fsgc often outperforms \psmap in generalization while significantly improving uncertainty quantification.
These results are presented in \Cref{appsec:real-world}.

\setlength{\tabcolsep}{1pt}
\begin{table}[ht]
\small
\caption{Starting from a pretrained checkpoint of ResNet18 on ImageNet \citep{Russakovsky2014ImageNetLS}, we report the performance on CIFAR-10 \citep{recht2018cifar10.1}. \fsgc benefits predictive performance and calibration. Means and standard errors are computed over five seeds.}
\label{tab:c10_transfer}
\vspace{-7pt}
\begin{adjustbox}{width=\linewidth}
\begin{sc}
\begin{tabular}{l|ccccc}
\toprule
Method & Acc. $\uparrow$ & Sel. Pred. $\uparrow$ & NLL $\downarrow$ & ECE $\downarrow$ & OOD $\uparrow$ \\
\midrule
\psmap & $96.2\% \pms{0.1}$ & $99.6\% \pms{0.0}$ & $0.13 \pms{0.01}$ & $3.2\% \pms{0.2}$ & $96.3\% \pms{0.7}$ \\
\fsgc & $96.2\% \pms{0.1}$ & $99.6\% \pms{0.0}$ & \cellcolor[gray]{0.9}$\mathbf{0.11} \pms {0.00}$ & \cellcolor[gray]{0.9}$\mathbf{1.3}\% \pms{0.1}$ & \cellcolor[gray]{0.9}$\mathbf{98.9}\% \pms{0.1}$ \\
\bottomrule
\end{tabular}
\end{sc}
\end{adjustbox}
\end{table}

%% file: icml2023/6_conclusion.tex
\section{Conclusion}

We presented a probabilistic perspective on function-space regularization in neural networks and used it to derive function-space empirical Bayes (\fsgc)---a method that combines parameter- and function-spaces regularization.
We demonstrated that \fsgc exhibits desirable empirical properties, such as significantly improved predictive uncertainty quantification both in-distribution and under semantic shift.
\fsgc is scalable, can be applied to any neural network architecture, can be used with pretrained models, and allows effectively incorporating prior information in a probabilistically principled manner.

%% file: icml2023/7_acknowledgements.tex
\section*{Acknowledgments}

We thank anonymous reviewers for useful feedback.
This work is supported by NSF CAREER IIS-2145492, NSF I-DISRE 193471, NIH R01DA048764-01A1, NSF IIS-1910266, NSF 1922658 NRT-HDR, Meta Core Data Science, Google AI Research, BigHat Biosciences, Capital One, and an Amazon Research Award.

%% file: icml2023/8_appendices.tex
\begin{appendices}

\crefalias{section}{appsec}
\crefalias{subsection}{appsec}
\crefalias{subsubsection}{appsec}

\setcounter{equation}{0}
\renewcommand{\theequation}{\thesection.\arabic{equation}}

\onecolumn

\vspace*{-20pt}

\section*{\LARGE \centering Appendix
}
\label{sec:appendix}

\vspace{0.2in}
{\hrule height 0.3mm}
\vspace{14pt}

\section{Additional Details and Experiments}
\label{appsec:experiments}

\subsection{Hyperparameters}

In \Cref{tab:hyper_ranges}, we provide the key hyperparameters used with \fsgc. We operate over the search space using randomized grid search. In addition to the learning rate $\eta$, cosine scheduler $\alpha$, and weight decay used by standard \psmap, we use two more hyperparameters---the prior variance $\tau^{-1}_{f}$ and the number of Monte Carlo samples $J$.

\setlength{\tabcolsep}{77pt}
\begin{table}[!ht]
\vspace*{-5pt}
    \centering
    \caption{Hyperparameter Ranges}
    \vspace*{-7pt}
    \begin{sc}
    \begin{tabular}{c|c}
    \toprule
       Hyperparameter  & Range  \\
       \midrule
       Learning Rate $\eta$ & $[10^{-10}, 10^{-1}]$ \\
       Scheduler $\alpha$ & $[0,1]$ \\
       Weight Decay $\tau^{-1}_{\theta}$ & $[10^{-10}, 1]$ \\
       Prior Variance $\tau^{-1}_{f}$ & $[10^{-7}, 5\times 10^{4}]$ \\
       Monte Carlo Samples $J$ & $\{1,2,5,10\}$ \\
       \bottomrule
    \end{tabular}
    \end{sc}
    \label{tab:hyper_ranges}
\vspace*{-5pt}
\end{table}

\subsection{Deep Ensembles}

\citet{lakshminarayanan2017simple} propose a simple alternative to Bayesian neural networks by computing the Bayesian model average using a set of independently trained neural networks, i.e. the softmax outputs from each independent network are averaged to provide the final predictive distribution for classification. This method is called Deep Ensembles.
Across literature, Deep Ensembles have been observed to provide improved generalization and better calibration. Subsequently, in \Cref{tab:ens_results}, we quantify the benefit of Deep Ensembles for \fsgc. Surprisingly, we find that Deep Ensembles benefit \psmap more than they do \fsgc.
A key property of ensemble components that lead to better generalization is the induced diversity \citep{Breiman2001RandomF}. 
We speculate that \fsgc may enforce a bias that makes the components of an ensemble less diverse, since it has a more informative prior than standard weight decay.

\setlength{\tabcolsep}{2.5pt}
\begin{table}[!ht]
\centering
\caption{We report the accuracy ({\sc acc.}), negative log-likelihood ({\sc nll}), expected calibration error ({\sc ece}), area under selective prediction accuracy curve ({\sc Sel. Pred.}), and area under OOD prediction accuracy curve ({\sc OOD}) for FashionMNIST \citep{xiao2017/online} and CIFAR-10 \citep{Krizhevsky2010ConvolutionalDB} with \fsgc deep ensembles \citep{lakshminarayanan2017simple}.}
\vspace*{-7pt}
\label{tab:ens_results}
\begin{sc}
\begin{tabular}{l|ccccc|ccccc}
\toprule
& \multicolumn{5}{c}{FashionMNIST} & \multicolumn{5}{c}{CIFAR-10} \\
\midrule
Method & Acc. $\uparrow$ & Sel. Pred. $\uparrow$ & NLL $\downarrow$ & ECE $\downarrow$ & OOD $\uparrow$ & Acc. $\uparrow$ & Sel. Pred. $\uparrow$ & NLL $\downarrow$ & ECE $\downarrow$ & OOD $\uparrow$ \\
\midrule
\psmap-\textsc{ensemble} & $94.5\%$ & $99.3\%$ & $\mathbf{0.18}$ & $\mathbf{1.6}\%$ & $94.9\%$  & $\mathbf{96.0}\%$ & $\mathbf{99.6}\%$ & $\mathbf{0.13}$ & $\mathbf{0.7}\%$ & $95.7\%$ \\
\fsgc-\textsc{ensemble} & $\mathbf{94.7}\%$ & $\mathbf{98.9}\%$ & $0.21$ & $3.7\%$ & $\mathbf{99.9}\%$ & $95.8\%$ & $99.5\%$ & $0.17$ & $3.0\%$ & $\mathbf{99.1}\%$  \\
\bottomrule
\end{tabular}
\end{sc}
\vspace*{-5pt}
\end{table}

\subsection{Performance with CIFAR-10.1}

\citet{recht2018cifar10.1} introduce an extended set of test samples similar in distribution to CIFAR-10 meant as a safeguard against overfitting of methods to benchmark classification task of CIFAR-10.
In \Cref{tab:c101}, we report the performance metrics for CIFAR-10 trained models evaluated on the CIFAR-10.1 test set.

\setlength{\tabcolsep}{24pt}
\begin{table}[!ht]
\centering
\caption{We report the accuracy ({\sc acc.}), negative log-likelihood ({\sc nll}), expected calibration error ({\sc ece}), area under selective prediction accuracy curve ({\sc Sel. Pred.}), and area under OOD prediction accuracy curve ({\sc OOD}) for CIFAR-10.1 \citep{recht2018cifar10.1} using models trained on CIFAR-10. Means and standard errors are computed over five seeds.}
\vspace*{-7pt}
\label{tab:c101}
\begin{sc}
\begin{tabular}{l|ccccc}
\toprule
Method & Acc. $\uparrow$ & Sel. Pred. $\uparrow$ & NLL $\downarrow$ & ECE $\downarrow$ \\
\midrule
\psmap & $\mathbf{88.0}\% \pm 0.1$ & $97.5\% \pm 0.1$ & $0.49 \pm 0.00$ & $7.6\% \pm 0.1$ \\
\fsgc & $86.8\% \pm 0.4$ & $97.2\% \pm 0.2$ & $0.49 \pm 0.01$ & $\mathbf{4.0}\% \pm 0.2$ \\
\bottomrule
\end{tabular}
\end{sc}
\vspace*{-10pt}
\end{table}

\clearpage

\subsection{Model Robustness with CIFAR-10 Corrupted}

\citet{hendrycks2018benchmarking} propose the CIFAR-10 Corrupted dataset as a test for model robustness, which consists of 19 commonly observed corruptions of images including blur, noise, and pixelation. 
All corruptions are created with CIFAR-10 test images at five different levels.

\begin{figure}[!ht]
    \centering
    \includegraphics[width=0.97\linewidth]{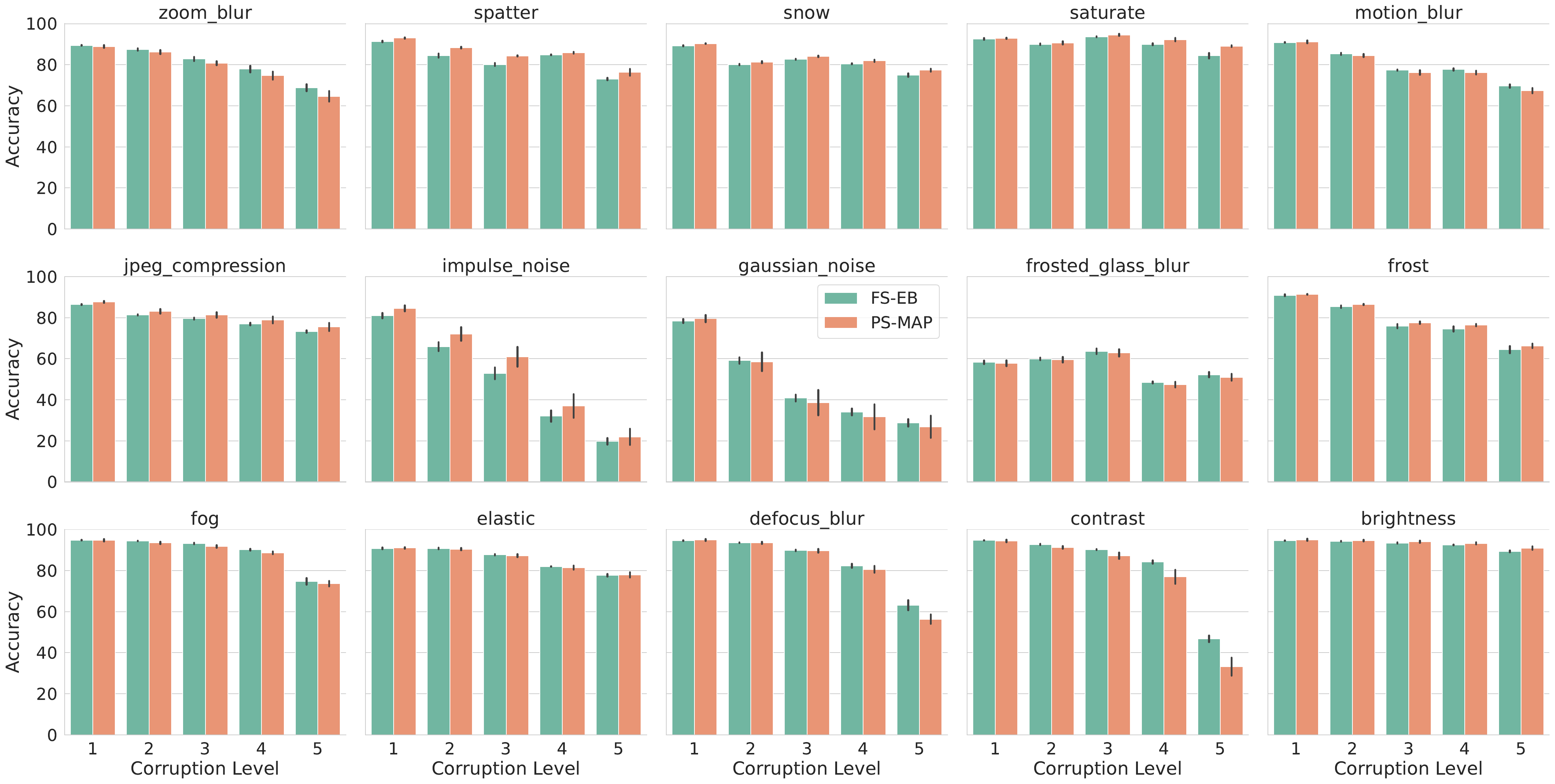}
    \caption{Accuracy on CIFAR-10 Corrupted}
    \label{fig:c10_acc_all}
\end{figure}

\begin{figure}[!ht]
    \centering
    \includegraphics[width=0.97\linewidth]{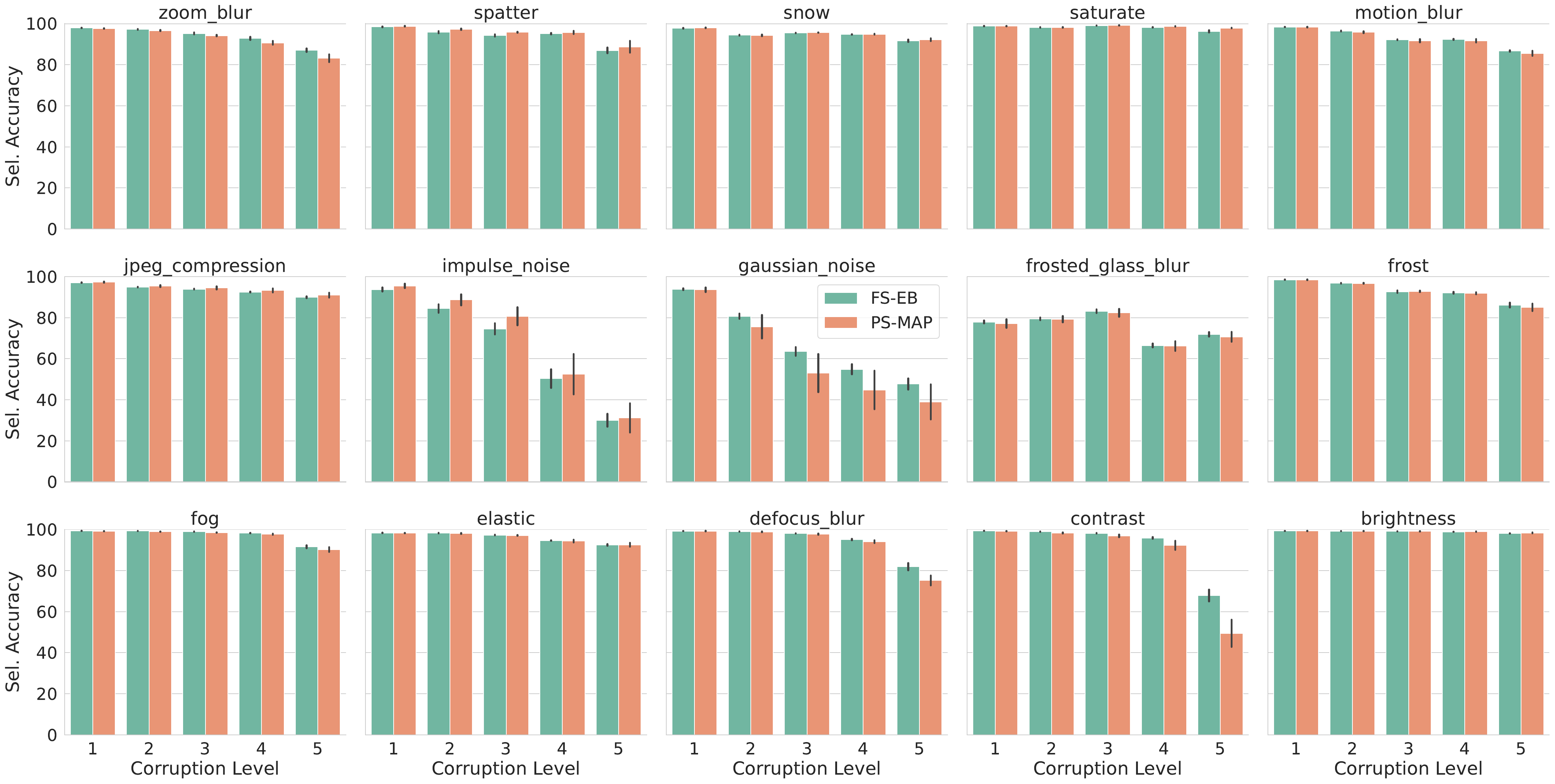}
    \caption{Selective Accuracy on CIFAR-10 Corrupted}
    \label{fig:c10_sel_acc_all}
\end{figure}

In continuation of the discussion around \Cref{fig:c10c_acc}, we summarize the accuracy and selective accuracy across all the corruptions in \Cref{fig:c10_acc_all,fig:c10_sel_acc_all}.

\clearpage

\subsection{Effect of Training Data Size}

In \Cref{tab:fmnist_dataset_size,tab:c10_dataset_size}, we quantify the performance of \fsgc in the low-data regime. For various fractions (${10\%, 25\%, 50\%, 75\%}$) of the full training dataset, we train both \psmap and \fsgc.
Across all metrics, we find that \fsgc overall tends to outperform \psmap significantly.

\setlength{\tabcolsep}{10.5pt}
\begin{table}[!ht]
\centering
\vspace*{-5pt}
\caption{We assess the performance of \fsgc in the low training data regime for FashionMNIST. Overall, we find that \fsgc tends to generalize significantly better under small data, similar to our findings for FashionMNIST in \Cref{tab:c10_dataset_size}. Means and standard errors are computed over five seeds.}
\vspace*{-7pt}
\label{tab:fmnist_dataset_size}
\begin{sc}
\begin{tabular}{l|l|ccccc}
\toprule
Fraction & Method & Acc. $\uparrow$ & Sel. Pred. $\uparrow$ & NLL $\downarrow$ & ECE $\downarrow$ & OOD AUROC $\uparrow$  \\
\midrule
\multirow{2}{*}{$10\%$} & \fsgc & $\mathbf{89.0}\% \pms{0.1}$ & $\mathbf{97.2}\% \pms{0.1}$ & $\mathbf{0.47} \pms{0.01}$ & $\mathbf{6.7}\% \pms{0.1}$ & $\mathbf{98.1}\% \pms{0.4}$ \\
& \psmap & $88.1\% \pms{0.2}$ & $97.0\% \pms{0.1}$ & $0.49 \pms{0.00}$ & $7.4\% \pms{0.1}$ & $88.1\% \pms{2.1}$ \\
\midrule
\multirow{2}{*}{$25\%$} & \fsgc & $\mathbf{91.5}\% \pms{0.1}$ & $98.0\% \pms{0.1}$ & $\mathbf{0.35} \pms{0.01}$ & $\mathbf{5.2}\% \pms{0.1}$ & $\mathbf{98.6}\% \pms{0.2}$ \\
& \psmap & $91.1\% \pms{0.1}$ & $\mathbf{98.3}\% \pms{0.0}$ & $0.36 \pms{0.00}$ & $5.4\% \pms{0.1}$ & $88.6\% \pms{1.3}$ \\
\midrule
\multirow{2}{*}{$50\%$} & \fsgc & $\mathbf{92.9}\% \pms{0.0}$ & $98.2\% \pms{0.1}$ & $0.31 \pms{0.00}$ & $\mathbf{4.6}\% \pms{0.1}$ & $\mathbf{99.5}\% \pms{0.1}$ \\
& \psmap & $92.5\% \pms{0.1}$ & $\mathbf{98.7}\% \pms{0.0}$ & $\mathbf{0.30} \pms{0.01}$ & $4.5\% \pms{0.1}$ & $93.0\% \pms{0.2}$ \\
\midrule
\multirow{2}{*}{$75\%$} & \fsgc & $\mathbf{93.6}\% \pms{0.1}$ & $98.3\% \pms{0.0}$ & $0.29 \pms{0.00}$ & $4.4\% \pms{0.1}$ & $\mathbf{99.8}\% \pms{0.0}$ \\
& \psmap & $93.2\% \pms{0.1}$ & $\mathbf{98.9}\% \pms{0.0}$ & $\mathbf{0.28} \pms{0.00}$ & $\mathbf{4.2}\% \pms{0.1}$ & $93.1\% \pms{0.7}$ \\
\midrule
\multirow{2}{*}{$100\%$} & \fsgc & $\mathbf{94.1}\% \pms{0.1}$ & $98.8\% \pms{0.0}$ & $\mathbf{0.19} \pms{0.00}$ & $\mathbf{1.8}\% \pms{0.1}$ & $\mathbf{99.9}\% \pms{0.0}$ \\
& \psmap & $93.8\% \pms{0.0}$ & $\mathbf{98.9}\% \pms{0.0}$ & $0.26 \pms{0.00}$ & $3.6\% \pms{0.0}$ & $94.9\% \pms{0.4}$ \\

& \psmap & $91.1\% \pms{0.1}$ & $\mathbf{98.3}\% \pms{0.0}$ & $0.36 \pms{0.00}$ & $5.4\% \pms{0.1}$ & $88.6\% \pms{1.3}$ \\
\bottomrule
\end{tabular}
\end{sc}
\vspace*{-8pt}
\end{table}

\setlength{\tabcolsep}{10.5pt}
\begin{table}[!ht]
\centering
\caption{We assess the performance of \fsgc in the low training data regime for CIFAR-10. Overall, we find that \fsgc tends to generalize significantly better under small data, similar to our findings for FashionMNIST in \Cref{tab:fmnist_dataset_size}. Means and standard errors are computed over five seeds.}
\vspace*{-7pt}
\label{tab:c10_dataset_size}
\begin{sc}
\begin{tabular}{l|l|ccccc}
\toprule
Fraction & Method & Acc. $\uparrow$ & Sel. Pred. $\uparrow$ & NLL $\downarrow$ & ECE $\downarrow$ & OOD AUROC $\uparrow$  \\
\midrule
\multirow{2}{*}{$10\%$} & \fsgc & $\mathbf{78.3}\% \pms{0.1}$ & $\mathbf{93.2}\% \pms{0.0}$ & $\mathbf{0.83} \pms{0.00}$ & $\mathbf{11.1}\% \pms{0.3}$ & $\mathbf{95.9}\% \pms{0.3}$ \\
& \psmap & $72.7\% \pms{0.1}$ & $89.9\% \pms{0.1}$ & $1.36 \pms{0.00}$ & $19.7\% \pms{0.0}$ & $66.2\% \pms{1.0}$ \\
\midrule
\multirow{2}{*}{$25\%$} & \fsgc & $\mathbf{87.6}\% \pms{0.0}$ & $\mathbf{97.2}\% \pms{0.0}$ & $\mathbf{0.47} \pms{0.00}$ & $\mathbf{6.0}\% \pms{0.1}$ & $\mathbf{99.6}\% \pms{0.0}$ \\
& \psmap & $87.1\% \pms{0.4}$ & $97.1\% \pms{0.1}$ & $0.54 \pms{0.01}$ &	$7.9\% \pms{0.2}$	& $74.8\% \pms{2.5}$ \\
\midrule
\multirow{2}{*}{$50\%$} & \fsgc & $92.0\% \pms{0.1}$ & $98.7\% \pms{0.0}$ & $\mathbf{0.30} \pms{0.00}$ & $\mathbf{2.6}\% \pms{0.1}$ & $\mathbf{99.9}\% \pms{0.0}$ \\
& \psmap & $\mathbf{92.5}\% \pms{0.0}$ & $98.7\% \pms{0.0}$ & $0.32 \pms{0.01}$ & $4.7\% \pms{0.1}$ & $85.9\% \pms{1.3}$ \\
\midrule
\multirow{2}{*}{$75\%$} & \fsgc & $93.9\% \pms{0.1}$ & $99.1\% \pms{0.0}$ & $0.23 \pms{0.0}$ & $\mathbf{1.8}\% \pms{0.0}$ & $\mathbf{99.9}\% \pms{0.0}$ \\
& \psmap & $\mathbf{94.4}\% \pms{0.0}$ & $99.1\% \pms{0.0}$ & $0.23 \pms{0.00}$ & $3.4\% \pms{0.0}$ & $91.6\% \pms{0.8}$ \\
\midrule
\multirow{2}{*}{$100\%$} & \fsgc & $\mathbf{95.1}\% \pms{0.1}$ & $\mathbf{99.4}\% \pms{0.0}$ & $\mathbf{0.20} \pms{0.00}$ & $\mathbf{2.1}\% \pms{0.1}$ & $\mathbf{99.4}\% \pms{0.0}$ \\
& \psmap & $94.9\% \pms{0.1}$ & $99.3\% \pms{0.0}$ & $0.21 \pms{0.01}$ & $3.0\% \pms{0.0}$ & $93.0\% \pms{0.2}$ \\
\bottomrule
\end{tabular}
\end{sc}
\end{table}

\subsection{Effect of Context Set Batch Size}

During each gradient step of \fsgc training, we use a subset of points from the context distribution, sampled uniformly at random as described in \Cref{sec:fsgc}. The number of samples is what we call the context set batch size. In \Cref{tab:ctx_size}, we vary this batch size and find that most metrics are not very sensitive to this hyperparameter choice.

\begin{table}[!ht]
\centering
\caption{We vary the size of the context set batch size ad assess the effect on predictive performance.}\vspace*{-7pt}
\label{tab:ctx_size}
\begin{adjustbox}{width=\linewidth}
\begin{sc}
\begin{tabular}{l|ccccc|ccccc}
\toprule
& \multicolumn{5}{c}{FashionMNIST} & \multicolumn{5}{c}{CIFAR-10} \\
\midrule
Batch Size & Acc. $\uparrow$ & Sel. Pred. $\uparrow$ & NLL $\downarrow$ & ECE $\downarrow$ & OOD $\uparrow$ & Acc. $\uparrow$ & Sel. Pred. $\uparrow$ & NLL $\downarrow$ & ECE $\downarrow$ & OOD $\uparrow$ \\
\midrule
32 & $94.1\% \pm 0.0$ & $\mathbf{98.4}\% \pm 0.1$ & $\mathbf{0.27} \pm 0.00$ & $\mathbf{4.1}\% \pm 0.0$ & $98.9\% \pm 0.1$ & $95.0\% \pm 0.1$ & $99.3\% \pm 0.0$ & $0.19 \pm 0.00$ & $1.5\% \pm 0.1$ & $99.9\% \pm 0.0$ \\
64 & $94.1\% \pm 0.0$ & $98.3\% \pm 0.0$ & $\mathbf{0.27} \pm 0.00$ & $\mathbf{4.1}\% \pm 0.0$ & $99.5\% \pm 0.0$ & $94.9\% \pm 0.1$ & $99.3\% \pm 0.0$ & $0.19 \pm 0.0$ & $\mathbf{1.4}\% \pm 0.0$ & $99.9\% \pm 0.0$ \\
128 & $94.1\% \pm 0.0$ & $98.3\% \pm 0.0$ & $0.28 \pm 0.00$ & $4.2\% \pm 0.0$ & $\mathbf{99.9}\% \pm 0.0$ & $\mathbf{95.1}\% \pm 0.1$ & $\mathbf{99.4}\% \pm 0.0$ & $0.20 \pm 0.00$ & $2.1\% \pm 0.1$ & $99.4\% \pm 0.0$ \\
\bottomrule
\end{tabular}
\end{sc}
\end{adjustbox}
\end{table}

\clearpage

\vspace*{-10pt}
\subsection{Effect of Training Context Distribution}

We study the effect of different context set distributions. 
In our main experiments, we use KMNIST \citep{Clanuwat2018DeepLF} as the context distribution for FashionMNIST and CIFAR-100 as the context distribution for CIFAR-10.
In \Cref{tab:ctx_dist}, we evaluate the performance of \fsgc with the context set being (i) the training inputs and (ii) corrupted training inputs.

\setlength{\tabcolsep}{1.0pt}
\begin{table}[!ht]
\centering
\vspace*{-5pt}
\caption{We vary the context set ({\sc ctx. set}) distribution to be (i) the training set, and (ii) the training set with data augmentations and quantify the performance of $\fsgc$. $\X_C$ = KMNIST for FashionMNIST and $\X_C$ = CIFAR-100 for CIFAR-10. Changing the context set distribution does have a significant impact on generalization performance in terms of accuracy and can also lead to significant improvement in out-of-distribution detection.}
\vspace*{-7pt}
\label{tab:ctx_dist}
\begin{adjustbox}{width=\linewidth}
\begin{sc}
\begin{tabular}{l|ccccc|ccccc}
\toprule
& \multicolumn{5}{c}{FashionMNIST} & \multicolumn{5}{c}{CIFAR-10} \\
\midrule
Ctx. Set & Acc. $\uparrow$ & Sel. Pred. $\uparrow$ & NLL $\downarrow$ & ECE $\downarrow$ & OOD $\uparrow$ & Acc. $\uparrow$ & Sel. Pred. $\uparrow$ & NLL $\downarrow$ & ECE $\downarrow$ & OOD $\uparrow$  \\
\midrule
Train & $93.9\% \pms{0.0}$ & $98.3\% \pms{0.1}$ & $0.28 \pms{0.00}$ & $4.2\% \pms{0.0}$ & $97.6\% \pms{0.5}$ & $94.9\% \pms{0.1}$ & $99.3\% \pms{0.0}$ & $\mathbf{0.19} \pms{0.00}$ & $1.7\% \pms{0.1}$ & $92.1\% \pms{0.6}$ \\
Train Corr. & $\mathbf{94.1}\% \pms{0.0}$ & $98.4\% \pms{0.0}$ & $0.27 \pms{0.00}$ & $4.1\% \pms{0.0}$ & $97.7\% \pms{0.5}$ & $94.7\% \pms{0.1}$ & $99.2\% \pms{0.0}$ & $0.20 \pms{0.00}$ & $\mathbf{1.4}\% \pms{0.0}$ & $99.9\% \pms{0.0}$ \\
$\X_C$ & $\mathbf{94.1}\% \pms{0.1}$ & $\mathbf{98.8}\% \pms{0.0}$ & $\mathbf{0.19} \pms{0.00}$ & $\mathbf{1.8}\% \pms{0.1}$ & $\mathbf{99.9}\% \pms{0.0}$ & $\mathbf{95.1}\% \pms{0.1}$ & $\mathbf{99.4}\% \pms{0.0}$ & $0.20 \pms{0.00}$ & $2.1\% \pms{0.1}$ & $\mathbf{99.4}\% \pms{0.1}$ \\
\bottomrule
\end{tabular}
\end{sc}
\end{adjustbox}
\vspace*{-5pt}
\end{table}

\subsection{Transfer Learning on Real-World Datasets}
\label{appsec:real-world}

In addition to standard benchmark datasets, we also consider three additional real-world datasets - APTOS Blindness Detection \citep{aptos,aptos1}, Melanoma Classification \citep{melanoma,melanoma1,melanoma2}, and Cassava Leaf Disease Classification \citep{mwebaze2019icassava,cassava1}

\setlength{\tabcolsep}{21pt}
\begin{table}[!ht]
\centering
\caption{Performance on Real-World Datasets, transfer learning from an ImageNet-pretrained ResNet-50 \citep{he2016deep}.}
\vspace*{-7pt}
\label{tab:real_world}
\begin{sc}
\begin{tabular}{l|l|cccc}
\toprule
Dataset & Method & Acc. $\uparrow$ & Sel. Pred. $\uparrow$ & NLL $\downarrow$ & ECE $\downarrow$ \\
\midrule
\multirow{2}{*}{APTOS} & \fsgc & $83.2\%$ & $94.2\%$ & $\mathbf{0.78}$ & $\mathbf{11.3}\%$ \\
& \psmap & $\mathbf{83.7}\%$ & $\mathbf{93.7}\%$ & $0.83$ & $12.8\%$ \\
\midrule
\multirow{2}{*}{Melanoma} & \fsgc & $\mathbf{98.6}\%$ & $\mathbf{99.8}\%$ & $\mathbf{0.05}$ & $\mathbf{1.6}\%$ \\
& \psmap & $98.2\%$ & $99.7\%$ & $0.08$ & $1.8\%$ \\
\midrule
\multirow{2}{*}{Cassava} & \fsgc & $86.5\%$ & $\mathbf{96.5}\%$ & $\mathbf{0.64}$ & $\mathbf{9.0}\%$ \\
& \psmap & $86.5\%$ & $95.6\%$ & $0.80$ & $10.9\%$ \\
\bottomrule
\end{tabular}
\end{sc}
\end{table}

Using an ImageNet-pretrained \citep{Russakovsky2014ImageNetLS} ResNet-50 \citep{he2016deep}, similar in spirit to \citet{Fang2023DoesPO}, we conduct a transfer learning experiment. In \Cref{tab:real_world}, we provide the performance of \fsgc on these datasets and find that \fsgc can often provide improvements in the data fit in terms of the data likelihood and much better calibration in terms of ECE \citep{Naeini2015ObtainingWC}.
\subsection{Runtimes}

For reference, we provide approximate runtimes of \fsgc and \psmap in \Cref{tab:runtimes}.

\setlength{\tabcolsep}{33pt}
\begin{table}[!ht]
    \centering
    \caption{Approximate runtime for a single gradient step and one full epoch of training for FashionMNIST and CIFAR-10.}
    \vspace*{-7pt}
    \label{tab:runtimes}
    \begin{tabular}{l|l|cc}
\toprule
Dataset & Method & Gradient Step (ms) $\downarrow$ & Epoch (s) $\uparrow$ \\
\midrule
\multirow{3}{*}{FashionMNIST} & \psmap & $40$ & $18$ \\
 & \fsgc & $129$ & $60$ \\
 & \fsvi & $319$ & $144$ \\
\midrule
\multirow{3}{*}{CIFAR-10} & \psmap & $55$ & $21$ \\
& \fsgc & $137$ & $61$ \\
 & \fsvi & $389$ & $189$ \\
\bottomrule
    \end{tabular}
\end{table}

\end{appendices}

%% file: main.bbl
\begin{thebibliography}{48}
\providecommand{\natexlab}[1]{#1}
\providecommand{\url}[1]{\texttt{#1}}
\expandafter\ifx\csname urlstyle\endcsname\relax
  \providecommand{\doi}[1]{doi: #1}\else
  \providecommand{\doi}{doi: \begingroup \urlstyle{rm}\Url}\fi

\bibitem[{Asia Pacific Tele-Ophthalmology Society}(2019)]{aptos}
{Asia Pacific Tele-Ophthalmology Society}.
\newblock Aptos 2019 blindness detection, 2019.
\newblock URL
  \url{https://www.kaggle.com/competitions/aptos2019-blindness-detection/overview}.

\bibitem[Band et~al.(2021)Band, Rudner, Feng, Filos, Nado, Dusenberry, Jerfel,
  Tran, and Gal]{Band2021benchmarking}
Band, N., Rudner, T. G.~J., Feng, Q., Filos, A., Nado, Z., Dusenberry, M.~W.,
  Jerfel, G., Tran, D., and Gal, Y.
\newblock {B}enchmarking {B}ayesian {D}eep {L}earning {o}n {D}iabetic
  {R}etinopathy {D}etection {T}asks.
\newblock In \emph{Advances in Neural Information Processing Systems 34}, 2021.

\bibitem[Benjamin et~al.(2018)Benjamin, Rolnick, and
  Kording]{Benjamin2018MeasuringAR}
Benjamin, A.~S., Rolnick, D., and Kording, K.~P.
\newblock Measuring and regularizing networks in function space.
\newblock \emph{ArXiv}, abs/1805.08289, 2018.

\bibitem[Bietti \& Mairal(2018)Bietti and Mairal]{bietti2018group}
Bietti, A. and Mairal, J.
\newblock Group invariance, stability to deformations, and complexity of deep
  convolutional representations, 2018.

\bibitem[Bietti et~al.(2019)Bietti, Mialon, Chen, and Mairal]{bietti2019kernel}
Bietti, A., Mialon, G., Chen, D., and Mairal, J.
\newblock A kernel perspective for regularizing deep neural networks.
\newblock In \emph{International Conference on Machine Learning}, pp.\
  664--674. PMLR, 2019.

\bibitem[Bishop(2006)]{Bishop2006PatternRA}
Bishop, C.~M.
\newblock Pattern recognition and machine learning (information science and
  statistics).
\newblock 2006.

\bibitem[Breiman(2001)]{Breiman2001RandomF}
Breiman, L.
\newblock Random forests.
\newblock \emph{Machine Learning}, 45:\penalty0 5--32, 2001.

\bibitem[Brown et~al.(2020)Brown, Mann, Ryder, Subbiah, Kaplan, Dhariwal,
  Neelakantan, Shyam, Sastry, Askell, Agarwal, Herbert-Voss, Krueger, Henighan,
  Child, Ramesh, Ziegler, Wu, Winter, Hesse, Chen, Sigler, Litwin, Gray, Chess,
  Clark, Berner, McCandlish, Radford, Sutskever, and
  Amodei]{Brown2020LanguageMA}
Brown, T.~B., Mann, B., Ryder, N., Subbiah, M., Kaplan, J., Dhariwal, P.,
  Neelakantan, A., Shyam, P., Sastry, G., Askell, A., Agarwal, S.,
  Herbert-Voss, A., Krueger, G., Henighan, T.~J., Child, R., Ramesh, A.,
  Ziegler, D.~M., Wu, J., Winter, C., Hesse, C., Chen, M., Sigler, E., Litwin,
  M., Gray, S., Chess, B., Clark, J., Berner, C., McCandlish, S., Radford, A.,
  Sutskever, I., and Amodei, D.
\newblock Language models are few-shot learners.
\newblock \emph{ArXiv}, abs/2005.14165, 2020.

\bibitem[Burt et~al.(2020)Burt, Ober, Garriga-Alonso, and van~der
  Wilk]{Burt2020UnderstandingVI}
Burt, D.~R., Ober, S., Garriga-Alonso, A., and van~der Wilk, M.
\newblock Understanding variational inference in function-space.
\newblock \emph{ArXiv}, abs/2011.09421, 2020.

\bibitem[Chen et~al.(2022)Chen, Shi, Rudner, Feng, Zhang, and
  Zhang]{chen2022ntk}
Chen, Z., Shi, X., Rudner, T. G.~J., Feng, Q., Zhang, W., and Zhang, T.
\newblock A neural tangent kernel perspective on function-space regularization
  in neural networks.
\newblock In \emph{OPT 2022: Optimization for Machine Learning (NeurIPS 2022
  Workshop)}, 2022.

\bibitem[Clanuwat et~al.(2018)Clanuwat, Bober-Irizar, Kitamoto, Lamb, Yamamoto,
  and Ha]{Clanuwat2018DeepLF}
Clanuwat, T., Bober-Irizar, M., Kitamoto, A., Lamb, A., Yamamoto, K., and Ha,
  D.
\newblock Deep learning for classical japanese literature.
\newblock \emph{ArXiv}, abs/1812.01718, 2018.

\bibitem[Denker \& LeCun(1990)Denker and LeCun]{Denker1990TransformingNO}
Denker, J.~S. and LeCun, Y.
\newblock Transforming neural-net output levels to probability distributions.
\newblock In \emph{NIPS}, 1990.

\bibitem[El-Yaniv \& Wiener(2010)El-Yaniv and Wiener]{elyaniv2010foundations}
El-Yaniv, R. and Wiener, Y.
\newblock On the foundations of noise-free selective classification.
\newblock \emph{Journal of Machine Learning Research}, 11\penalty0
  (53):\penalty0 1605--1641, 2010.

\bibitem[Fang et~al.(2023)Fang, Kornblith, and Schmidt]{Fang2023DoesPO}
Fang, A., Kornblith, S., and Schmidt, L.
\newblock Does progress on imagenet transfer to real-world datasets?
\newblock \emph{ArXiv}, abs/2301.04644, 2023.

\bibitem[Ha et~al.(2020)Ha, Liu, and Liu]{melanoma1}
Ha, Q., Liu, B., and Liu, F.
\newblock Identifying melanoma images using efficientnet ensemble: Winning
  solution to the {SIIM-ISIC} melanoma classification challenge.
\newblock \emph{CoRR}, abs/2010.05351, 2020.

\bibitem[Hanke(2021)]{cassava1}
Hanke, J.
\newblock 1st place solution, 2021.
\newblock URL
  \url{https://www.kaggle.com/competitions/cassava-leaf-disease-classification/discussion/221957}.

\bibitem[He et~al.(2016)He, Zhang, Ren, and Sun]{he2016deep}
He, K., Zhang, X., Ren, S., and Sun, J.
\newblock Deep residual learning for image recognition.
\newblock In \emph{2016 {IEEE} Conference on Computer Vision and Pattern
  Recognition, {CVPR} 2016, Las Vegas, NV, USA, June 27-30, 2016}, pp.\
  770--778. {IEEE} Computer Society, 2016.

\bibitem[Hendrycks \& Dietterich(2019)Hendrycks and
  Dietterich]{hendrycks2018benchmarking}
Hendrycks, D. and Dietterich, T.
\newblock Benchmarking neural network robustness to common corruptions and
  perturbations.
\newblock In \emph{International Conference on Learning Representations}, 2019.

\bibitem[Joo \& Chung(2020)Joo and Chung]{Joo2020DeepLR}
Joo, T. and Chung, U.
\newblock Revisiting explicit regularization in neural networks for
  well-calibrated predictive uncertainty, 2020.

\bibitem[Krizhevsky(2010)]{Krizhevsky2010ConvolutionalDB}
Krizhevsky, A.
\newblock Convolutional deep belief networks on cifar-10.
\newblock 2010.

\bibitem[Krogh \& Hertz(1991)Krogh and Hertz]{Krogh1991ASW}
Krogh, A. and Hertz, J.~A.
\newblock A simple weight decay can improve generalization.
\newblock In \emph{NIPS}, 1991.

\bibitem[Lakshminarayanan et~al.(2017)Lakshminarayanan, Pritzel, and
  Blundell]{lakshminarayanan2017simple}
Lakshminarayanan, B., Pritzel, A., and Blundell, C.
\newblock Simple and scalable predictive uncertainty estimation using deep
  ensembles.
\newblock In Guyon, I., von Luxburg, U., Bengio, S., Wallach, H.~M., Fergus,
  R., Vishwanathan, S. V.~N., and Garnett, R. (eds.), \emph{Advances in Neural
  Information Processing Systems 30: Annual Conference on Neural Information
  Processing Systems 2017, December 4-9, 2017, Long Beach, CA, {USA}}, pp.\
  6402--6413, 2017.

\bibitem[Ma \& Hern{\'a}ndez-Lobato(2021)Ma and
  Hern{\'a}ndez-Lobato]{Ma2021FunctionalVI}
Ma, C. and Hern{\'a}ndez-Lobato, J.~M.
\newblock Functional variational inference based on stochastic process
  generators.
\newblock In \emph{NeurIPS}, 2021.

\bibitem[Ma et~al.(2018)Ma, Li, and Hern{\'a}ndez-Lobato]{Ma2018VariationalIP}
Ma, C., Li, Y., and Hern{\'a}ndez-Lobato, J.~M.
\newblock Variational implicit processes.
\newblock In \emph{International Conference on Machine Learning}, 2018.

\bibitem[Murphy(2013)]{murphy2013probabilistic}
Murphy, K.~P.
\newblock \emph{Machine learning : a probabilistic perspective}.
\newblock MIT Press, Cambridge, Mass. [u.a.], 2013.
\newblock ISBN 9780262018029 0262018020.

\bibitem[Mwebaze et~al.(2019)Mwebaze, Gebru, Frome, Nsumba, and
  Tusubira]{mwebaze2019icassava}
Mwebaze, E., Gebru, T., Frome, A., Nsumba, S., and Tusubira, J.
\newblock icassava 2019fine-grained visual categorization challenge, 2019.

\bibitem[Nado et~al.(2021)Nado, Band, Collier, Djolonga, Dusenberry, Farquhar,
  Filos, Havasi, Jenatton, Jerfel, Liu, Mariet, Nixon, Padhy, Ren, Rudner, Wen,
  Wenzel, Murphy, Sculley, Lakshminarayanan, Snoek, Gal, and
  Tran]{Nado2021UncertaintyBaselines}
Nado, Z., Band, N., Collier, M., Djolonga, J., Dusenberry, M.~W., Farquhar, S.,
  Filos, A., Havasi, M., Jenatton, R., Jerfel, G., Liu, J., Mariet, Z., Nixon,
  J., Padhy, S., Ren, J., Rudner, T. G.~J., Wen, Y., Wenzel, F., Murphy, K.,
  Sculley, D., Lakshminarayanan, B., Snoek, J., Gal, Y., and Tran, D.
\newblock {U}ncertainty {B}aselines: {B}enchmarks {f}or {U}ncertainty {\&}
  {R}obustness {i}n {D}eep {L}earning.
\newblock 2021.

\bibitem[Naeini et~al.(2015)Naeini, Cooper, and
  Hauskrecht]{Naeini2015ObtainingWC}
Naeini, M.~P., Cooper, G.~F., and Hauskrecht, M.
\newblock Obtaining well calibrated probabilities using bayesian binning.
\newblock \emph{Proceedings of the ... AAAI Conference on Artificial
  Intelligence. AAAI Conference on Artificial Intelligence}, 2015:\penalty0
  2901--2907, 2015.

\bibitem[Netzer et~al.(2011)Netzer, Wang, Coates, Bissacco, Wu, and
  Ng]{Netzer2011ReadingDI}
Netzer, Y., Wang, T., Coates, A., Bissacco, A., Wu, B., and Ng, A.
\newblock Reading digits in natural images with unsupervised feature learning.
\newblock 2011.

\bibitem[Ober \& Aitchison(2020)Ober and Aitchison]{Ober2020GlobalIP}
Ober, S. and Aitchison, L.
\newblock Global inducing point variational posteriors for bayesian neural
  networks and deep gaussian processes.
\newblock In \emph{International Conference on Machine Learning}, 2020.

\bibitem[Pan(2020)]{melanoma2}
Pan, I.
\newblock [2nd place] solution overview, 2020.
\newblock URL
  \url{https://www.kaggle.com/competitions/siim-isic-melanoma-classification/discussion/175324}.

\bibitem[Rabanser et~al.(2022)Rabanser, Thudi, Hamidieh, Dziedzic, and
  Papernot]{rabanser2022selective}
Rabanser, S., Thudi, A., Hamidieh, K., Dziedzic, A., and Papernot, N.
\newblock Selective classification via neural network training dynamics, 2022.

\bibitem[Radford et~al.(2021)Radford, Kim, Hallacy, Ramesh, Goh, Agarwal,
  Sastry, Askell, Mishkin, Clark, Krueger, and
  Sutskever]{Radford2021LearningTV}
Radford, A., Kim, J.~W., Hallacy, C., Ramesh, A., Goh, G., Agarwal, S., Sastry,
  G., Askell, A., Mishkin, P., Clark, J., Krueger, G., and Sutskever, I.
\newblock Learning transferable visual models from natural language
  supervision.
\newblock In \emph{International Conference on Machine Learning}, 2021.

\bibitem[Recht et~al.(2018)Recht, Roelofs, Schmidt, and
  Shankar]{recht2018cifar10.1}
Recht, B., Roelofs, R., Schmidt, L., and Shankar, V.
\newblock Do cifar-10 classifiers generalize to cifar-10?
\newblock 2018.

\bibitem[Rudner et~al.(2022{\natexlab{a}})Rudner, Chen, Teh, and
  Gal]{rudner2022tractable}
Rudner, T. G.~J., Chen, Z., Teh, Y.~W., and Gal, Y.
\newblock Tractable function-space variational inference in {B}ayesian neural
  networks.
\newblock In Oh, A.~H., Agarwal, A., Belgrave, D., and Cho, K. (eds.),
  \emph{Advances in Neural Information Processing Systems}, 2022{\natexlab{a}}.

\bibitem[Rudner et~al.(2022{\natexlab{b}})Rudner, Smith, Feng, Teh, and
  Gal]{Rudner2022sfsvi}
Rudner, T. G.~J., Smith, F.~B., Feng, Q., Teh, Y.~W., and Gal, Y.
\newblock {C}ontinual {L}earning via {S}equential {F}unction-{S}pace
  {V}ariational {I}nference.
\newblock In \emph{Proceedings of the 38th International Conference on Machine
  Learning}, Proceedings of Machine Learning Research. PMLR,
  2022{\natexlab{b}}.

\bibitem[Russakovsky et~al.(2014)Russakovsky, Deng, Su, Krause, Satheesh, Ma,
  Huang, Karpathy, Khosla, Bernstein, Berg, and
  Fei-Fei]{Russakovsky2014ImageNetLS}
Russakovsky, O., Deng, J., Su, H., Krause, J., Satheesh, S., Ma, S., Huang, Z.,
  Karpathy, A., Khosla, A., Bernstein, M.~S., Berg, A.~C., and Fei-Fei, L.
\newblock Imagenet large scale visual recognition challenge.
\newblock \emph{International Journal of Computer Vision}, 115:\penalty0
  211--252, 2014.

\bibitem[SIIM \& ISIC(2020)SIIM and ISIC]{melanoma}
SIIM and ISIC.
\newblock Siim-isic melanoma classification, 2020.
\newblock URL
  \url{https://www.kaggle.com/competitions/siim-isic-melanoma-classification/overview}.

\bibitem[Sun et~al.(2019{\natexlab{a}})Sun, Zhang, Shi, and
  Grosse]{Sun2019FunctionalVB}
Sun, S., Zhang, G., Shi, J., and Grosse, R.~B.
\newblock Functional variational bayesian neural networks.
\newblock \emph{ArXiv}, abs/1903.05779, 2019{\natexlab{a}}.

\bibitem[Sun et~al.(2019{\natexlab{b}})Sun, Zhang, Shi, and
  Grosse]{sun2019fbnn}
Sun, S., Zhang, G., Shi, J., and Grosse, R.~B.
\newblock Functional variational {B}ayesian neural networks.
\newblock In \emph{7th International Conference on Learning Representations,
  {ICLR} 2019, New Orleans, LA, USA, May 6-9, 2019}. OpenReview.net,
  2019{\natexlab{b}}.

\bibitem[Titsias et~al.(2019)Titsias, Schwarz, de~G.~Matthews, Pascanu, and
  Teh]{Titsias2019FunctionalRF}
Titsias, M.~K., Schwarz, J., de~G.~Matthews, A.~G., Pascanu, R., and Teh, Y.~W.
\newblock Functional regularisation for continual learning using gaussian
  processes.
\newblock \emph{ArXiv}, abs/1901.11356, 2019.

\bibitem[Touvron et~al.(2023)Touvron, Lavril, Izacard, Martinet, Lachaux,
  Lacroix, Rozi{\`e}re, Goyal, Hambro, Azhar, Rodriguez, Joulin, Grave, and
  Lample]{Touvron2023LLaMAOA}
Touvron, H., Lavril, T., Izacard, G., Martinet, X., Lachaux, M.-A., Lacroix,
  T., Rozi{\`e}re, B., Goyal, N., Hambro, E., Azhar, F., Rodriguez, A., Joulin,
  A., Grave, E., and Lample, G.
\newblock Llama: Open and efficient foundation language models.
\newblock \emph{ArXiv}, abs/2302.13971, 2023.

\bibitem[Tran et~al.(2022)Tran, Liu, Dusenberry, Phan, Collier, Ren, Han, Wang,
  Mariet, Hu, Band, Rudner, Singhal, Nado, van Amersfoort~andAndreas Kirsch,
  Jenatton, Thain, Yuan, Buchanan, Murphy, Sculley, Gal, Ghahramani, Snoek, and
  Lakshminarayanan]{Tran2021plex}
Tran, D., Liu, J., Dusenberry, M.~W., Phan, D., Collier, M., Ren, J., Han, K.,
  Wang, Z., Mariet, Z., Hu, H., Band, N., Rudner, T. G.~J., Singhal, K., Nado,
  Z., van Amersfoort~andAndreas Kirsch, J., Jenatton, R., Thain, N., Yuan, H.,
  Buchanan, K., Murphy, K., Sculley, D., Gal, Y., Ghahramani, Z., Snoek, J.,
  and Lakshminarayanan, B.
\newblock {P}lex: {T}owards {R}eliability {U}sing {P}retrained {L}arge {M}odel
  {E}xtensions.
\newblock In \emph{ICML 2022 Workshop on Pre-training: Perspectives, Pitfalls,
  and Paths Forward}, 2022.

\bibitem[Wang et~al.(2019)Wang, Ren, Zhu, and Zhang]{Wang2019FunctionSP}
Wang, Z., Ren, T., Zhu, J., and Zhang, B.
\newblock Function space particle optimization for bayesian neural networks.
\newblock \emph{ArXiv}, abs/1902.09754, 2019.

\bibitem[Wilson \& Izmailov(2020)Wilson and Izmailov]{wilson2020Bayesian}
Wilson, A.~G. and Izmailov, P.
\newblock {B}ayesian deep learning and a probabilistic perspective of
  generalization.
\newblock In Larochelle, H., Ranzato, M., Hadsell, R., Balcan, M., and Lin, H.
  (eds.), \emph{Advances in Neural Information Processing Systems 33: Annual
  Conference on Neural Information Processing Systems 2020, NeurIPS 2020,
  December 6-12, 2020, virtual}, 2020.

\bibitem[Wolpert(1993)]{wolpert1993fsmap}
Wolpert, D.~H.
\newblock Bayesian backpropagation over i-o functions rather than weights.
\newblock In Cowan, J., Tesauro, G., and Alspector, J. (eds.), \emph{Advances
  in Neural Information Processing Systems}, volume~6. Morgan-Kaufmann, 1993.

\bibitem[Xiao et~al.(2017)Xiao, Rasul, and Vollgraf]{xiao2017/online}
Xiao, H., Rasul, K., and Vollgraf, R.
\newblock Fashion-mnist: a novel image dataset for benchmarking machine
  learning algorithms.
\newblock 2017.

\bibitem[Xu(2019)]{aptos1}
Xu, G.
\newblock 1st place solution summary, 2019.
\newblock URL
  \url{https://www.kaggle.com/competitions/aptos2019-blindness-detection/discussion/108065}.

\end{thebibliography}
